\documentclass{article}

\PassOptionsToPackage{numbers, compress}{natbib}
\usepackage[final]{neurips_data_2024}

\usepackage[utf8]{inputenc} 
\usepackage[T1]{fontenc}    
\usepackage{hyperref}       
\usepackage{url}            
\usepackage{booktabs}       
\usepackage{amsfonts}       
\usepackage{nicefrac}       
\usepackage{microtype}      
\usepackage{xcolor}         

\usepackage{xspace}
\usepackage{graphicx}
\usepackage{multirow}
\usepackage{tcolorbox}
\usepackage{amsmath}
\usepackage{amssymb}
\usepackage[normalem]{ulem}
\usepackage{colortbl}
\usepackage{utfsym}
\usepackage{pifont}
\usepackage{wasysym}
\usepackage{makecell}
\usepackage{fontawesome}
\usepackage{threeparttable}
\usepackage{enumitem}
\usepackage{wrapfig}

\def\eg{\textit{e.g.,} }
\def\ie{\textit{i.e.,} }
\def\bench{EvoCodeBench\xspace}

\useunder{\uline}{\ul}{}

\title{\bench: An Evolving Code Generation Benchmark with Domain-Specific Evaluations}

\author{%
  Jia Li \faMars$^{1,2}$, Ge Li$^{1,2}$\footnotemark[1], Xuanming Zhang$^3$, Yunfei Zhao$^{1,2}$, Yihong Dong$^{1,2}$, Zhi Jin$^{1,2}$ \\
  \textbf{Binhua Li$^4$, Fei Huang$^4$, Yongbin Li$^4$\footnotemark[1]} \\
  $^1$Key Laboratory of High Confidence Software Technologies (Peking University), Ministry of Education \\
  $^2$School of Computer Science, Peking University, Beijing, China \\
  $^3$Bytedance, $^4$Alibaba Group \\
  \texttt{lijia@stu.pku.edu.cn, lige@pku.edu.cn,  shuide.lyb@alibaba-inc.com} \\
}

\begin{document}

\maketitle

\renewcommand{\thefootnote}{\fnsymbol{footnote}}
\footnotetext[1]{Corresponding authors}
\renewcommand{\thefootnote}{\arabic{footnote}}

\begin{abstract}
How to evaluate Large Language Models (LLMs) in code generation remains an open question. 
Many benchmarks have been proposed, but they have two limitations, \ie data leakage and lack of domain-specific evaluation.
The former hurts the fairness of benchmarks, and the latter hinders practitioners from selecting superior LLMs for specific programming domains.

To address these two limitations, we propose a new benchmark - \textbf{\bench}, which has the following advances: 
\ding{182} \textbf{Evolving data.} \bench will be dynamically updated every period (\eg 6 months) to avoid data leakage. This paper releases the first version - \bench-2403, containing 275 samples from 25 repositories.
\ding{183} \textbf{A domain taxonomy and domain labels.} Based on the statistics of open-source communities, we design a programming domain taxonomy consisting of 10 popular domains. Based on the taxonomy, we annotate each sample in \bench with a domain label. \bench provides a broad platform for domain-specific evaluations.
\ding{184} \textbf{Domain-specific evaluations.}
Besides the Pass@$k$, we compute the Domain-Specific Improvement (DSI) and define LLMs' comfort and strange domains. These evaluations help practitioners select superior LLMs in specific domains and discover the shortcomings of existing LLMs.
Besides, \bench is collected by a rigorous pipeline and aligns with real-world repositories in multiple aspects (\eg code distributions).
We evaluate 8 popular LLMs (\eg gpt-4, DeepSeek Coder, StarCoder 2) on \bench and summarize some insights. \bench reveals the actual abilities of these LLMs in real-world repositories. For example, \textbf{the highest Pass@1 of gpt-4 on \bench-2403 is only 20.74\%.} Besides, we evaluate LLMs in different domains and discover their comfort and strange domains. For example, \textbf{gpt-4 performs best in most domains but falls behind others in the Internet domain. StarCoder 2-15B unexpectedly performs well in the Database domain and even outperforms 33B LLMs.} We release \bench, all prompts, and LLMs' completions for further community analysis\footnote{\url{https://github.com/seketeam/EvoCodeBench}}.

\end{abstract}

\section{Introduction}
\label{sec:Introduction}

Large Language Models (LLMs) have shown impressive abilities in code generation \cite{aiXcoder-7B,SCoT,AceCoder}. As more and more LLMs emerge, a reliable code generation benchmark is crucial to evaluating and selecting superior LLMs.
Many benchmarks have been proposed, such as HumanEval \cite{Codex}, ClassEval \cite{ClassEval}, and DevEval \cite{DevEval}. Researchers spend lots of effort to annotate test data manually and construct these benchmarks. For example, ClassEval and DevEval cost 500 and 674 person-hours, respectively.

Although promising, existing benchmarks have two limitations. 

\ding{182} \textbf{Data leakage (aka data contamination).} It means that test data is included in the training data. The trained models perform much better on leaked benchmarks than on real-world tasks. Because the training data of LLMs contains almost all open-source code repositories, existing benchmarks probably have data leakages \cite{CDD}. 
Researchers have to spend more effort to construct new benchmarks. 

\ding{183} \textbf{Lack of domain-specific evaluation.} Programming is highly domain-specific. Developers typically focus on specific domains (\eg database). Compared to comprehensive coding abilities, developers are more concerned about the performance of LLMs in specific domains. However, existing benchmarks lack domain labels or fall into narrow domains. Besides, they ignore domain-specific evaluations and analyses. Thus, the performance of LLMs across domains is still unclear.

\textbf{To alleviate the above limitations, we propose a new code generation benchmark named \bench.} \bench has three novelties. 
\ding{182} \textbf{Evolving data.} To avoid data leakages, \bench is an evolving benchmark and will be dynamically updated every period (\eg 6 months). Specifically, we build an automatic collection pipeline, which constructs new versions of \bench from the latest repositories. More details about the pipeline are in Section \ref{sec:bench:collection_pipeline}.
\ding{183} \textbf{A domain taxonomy and domain labels.} PyPI \cite{PyPI} is a popular open-source community containing code repositories from various domains. Based on the statistics of repositories on PyPI, we design a programming domain taxonomy covering 10 popular domains. Based on the taxonomy, we annotate each sample in \bench with a domain label. In the future, we will refine the taxonomy (\eg adding emerging domains) and provide a broad platform for domain-specific evaluations.
\ding{184} \textbf{Domain-specific evaluations.} Besides the Pass@$k$, we propose the Domain-Specific Improvement (DSI), which reflects the position of an LLM in specific domains. Based on the DSI, we define the comfort domains (\eg DSI > 10\%) and strange domains (\eg DSI < -10\%) of LLMs. These metrics allow practitioners to effectively select superior LLMs in specific domains. Model trainers can also discover which domains LLMs are weak to facilitate future iterations.

Besides the above advances, \bench has an advantage in data quality. 
\ding{185} \bench is collected from high-quality open-source repositories. More importantly, \bench aligns with real-world repositories in multiple aspects, \eg code distributions and dependency distributions. This ensures that the performance of LLMs on \bench reflects their abilities in real-world development scenarios.
\ding{186} \bench offers comprehensive annotations, \eg natural language requirements, original repositories, reference code, reference dependencies, domain labels, and test cases. \bench computes Pass@$k$ and Recall@$k$ to measure the correctness of generated programs in functionality and dependencies. 

In this paper, we release the first version - \bench-2403, which consists of 275 samples from 25 real-world repositories. Based on \bench-2403, we evaluate 8 popular LLMs (\ie gpt-4 \cite{gpt-4}, gpt-3.5 \cite{gpt-3.5}, DeepSeek Coder \cite{DeepSeek_Coder}, StarCoder 2 \cite{StarCoder-2}, CodeLLaMa \cite{CodeLLaMa}). 
Based on extensive experiments, we obtain the following insights. 
\ding{182} \bench significantly alleviates the data leakage and decreases the potential leak rate from 41.47\% to 2.18\%. 
\ding{183} \bench provides a reliable evaluation for repo-level code generation. We analyze these LLMs' failed cases and summarize future directions, \eg long context modeling 
\ding{184} We evaluate LLMs in different domains and discover their comfort domains and strange domains. For example, gpt-4 performs best in most domains but performs worse than others in the Internet domain. StarCoder 2-15B unexpectedly performs well in the Database domain and even outperforms 33B LLMs.

\begin{figure}[t]
\centering
\includegraphics[width=\textwidth]{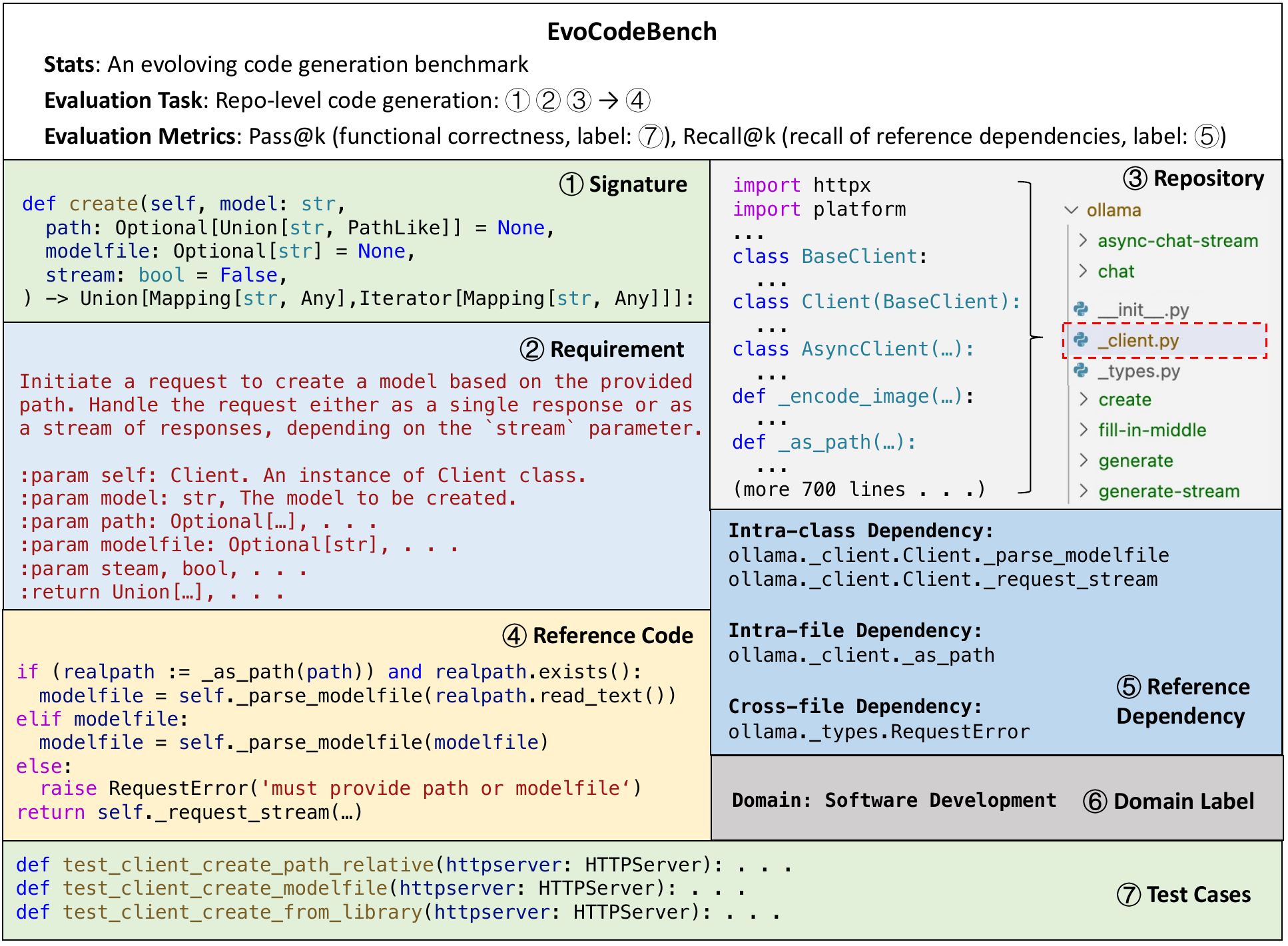}
\caption{An overview of \bench. Each sample consists of seven components.}
\label{fig:Benchmark_example}
\end{figure}

\section{\bench}
\label{sec:bench}

In this section, we first show an overview of \bench and then describe its tasks and evaluation metrics. Then, we present the first version - \bench-2403 and its statistics. Finally, we describe the automatic pipeline for constructing \bench.

\subsection{Overview}
\label{sec:bench:overview}

Figure \ref{fig:Benchmark_example} shows a sample in \bench. Each sample consists of seven components. 

\textbf{\ding{182} Function Signature:} The signature of the target code. 
\textbf{\ding{183} Requirement:} An English description detailing the functionality of the target code. 
\textbf{\ding{184} Repository:} The current repository contains hundreds of code files.
\textbf{\ding{185} Reference Code:} A developer-written implementation of the target code. This code may invoke dependencies defined in the current repository.
\textbf{\ding{186} Reference Dependency:} The dependencies invoked in the reference code include intra-class, intra-file, and cross-file dependencies.
\textbf{\ding{187} Domain Label:} The domain of the target code.
\textbf{\ding{188} Test Cases:} Test cases are used to check the functional correctness of the code.

\subsection{Task and Metrics}
\label{sec:bench:task_metric}

\bench evaluates LLMs in \textbf{repo-level code generation}. This task simulates the developers' coding process in a working repository. Given a requirement and a repository, LLMs are tasked to generate the code for the repository. Following previous work \cite{DevEval}, \bench contains two evaluation metrics, \ie Pass@$k$ and Recall@$k$.

\textbf{Pass@$k$ (Functional Correctness).} Following previous studies \cite{Codex,MBPP,CoderEval}, we assess the functional correctness of programs by executing test cases and compute the unbiased Pass@$k$.
Specifically, we generate $n \geq k$ programs per requirement, count the number of correct programs $c \leq n$ that pass test cases, and calculate the Pass@$k$:
\begin{equation}
\text{Pass}@k:=\underset{\text { Requirements }}{\mathbb{E}}\left[1-\frac{\left(\begin{array}{c}
n-c \\
k
\end{array}\right)}{\left(\begin{array}{l}
n \\
k
\end{array}\right)}\right]
\end{equation}

\textbf{Recall@$k$ (Recall of Reference Dependency).} 
We expect that generated programs can invoke relevant dependencies defined in contexts. Following previous work \cite{DevEval}, we report Recall@$k$, which computes the recall of reference dependencies in generated programs.   

Specifically, LLMs generate $k$ programs per requirement. The dependencies invoked by the $i$-th generated program are denoted as $\mathbb{P}_i$. We compare $\mathbb{P}_i$ with reference dependencies $\mathbb{R}$ and compute the Recall@$k$:
\begin{equation}
\text{Recall}@k:= \underset{\text{Requirements}}{\mathbb{E}} \left[ \max_{i \in [1, k]} \frac{|\mathbb{R} \cap \mathbb{P}_i|}{|\mathbb{R}|}\right]
\end{equation}
where $|\cdot|$ means the number of elements of a set.

\subsection{Benchmark Collection Pipeline}
\label{sec:bench:collection_pipeline}

We build an automatic pipeline for collecting \bench from the latest repositories. The pipeline consists of four stages as follows.

\textbf{Stage I: Repository selection and function scraping.} We crawl high-quality repositories from GitHub, satisfying the following criteria: open-source Python repositories with permissive licenses,
created within the last six months, non-fork and non-malicious projects, more than 50 stars, and having explicit unit tests. Then, we extract candidate functions from repositories and exclude trivial functions (\eg empty or initialization functions). 

\textbf{Stage II: Execution-based filtering.} For each candidate function, we extract test cases invoking it from current repositories. We use \texttt{pip} \cite{Pip} to install execution environments and leverage \texttt{Pytest} \cite{Pytest} to run test cases. Candidate functions without executable test cases are excluded. 

\textbf{Stage III: Automatic annotations.} We leverage a static analysis-based parser \cite{Pyan} to extract each candidate function's signature, function body (\ie reference code), and invoked dependencies (\ie reference dependencies). Because manually writing requirements is laborious, we use LLMs to generate requirements. Specifically, we craft a one-shot prompt, which teaches LLMs to write requirements in a specific format (\ie functional descriptions and input-output parameters). The prompt template is in Appendix \ref{sec:appendix:experiment:prompt}.

Next, we annotate each sample's domain label. To standardize the domains, we manually design a domain taxonomy. Specifically, we collect the statistics (\eg stars and domains) of repositories in a popular software community - PyPI \cite{PyPI}. Based on the statistics, we determine the top 10 domains with the most high-star repositories and construct the taxonomy. The 10 domains cover most of the repositories on PyPI and are shown in Table \ref{tab:domain_distribution}. In the future, we will continuously refine the taxonomy (\eg adding emerging domains).
Finally, we make a prompt and leverage LLMs to automatically annotate domain labels based on candidate functions and our taxonomy. Functions that do not satisfy any of the domains in our taxonomy are excluded. The prompt template is in Appendix \ref{sec:appendix:experiment:prompt}.

In Section \ref{sec:discussion}, we conduct a human evaluation to assess auto-generated annotations. The results show that auto-generated annotations are comparable to human-written ones in most cases (\ie requirement: 96.7\% samples and domain label: 98.5\% samples).

\begin{wraptable}{r}{0.3\linewidth}
\vspace{-10mm}
\centering
\caption{The domain distribution of \bench-2403.}
\label{tab:domain_distribution}
\resizebox{\linewidth}{!}{
\begin{tabular}{lc}
\toprule
Domain & Count \\ \midrule
Scientific Engineering & 120 \\
Software Development & 50 \\
Multimedia & 32 \\
Database & 18 \\
System & 17 \\
Internet & 15 \\
Text Processing & 12 \\
Communications & 8 \\
Utilities & 2 \\
Security & 1 \\
\bottomrule
\end{tabular}}
\end{wraptable}

\textbf{Stage IV: Benchmark Construction.} Finally, we randomly select several candidate functions to construct \bench. Following the related work \cite{DevEval}, we strive to make \bench satisfy the following goals: consistent with the code distribution observed in 500 real-world repositories, close to the average number of dependencies in 500 real-world repositories, including as many samples as possible. We have anonymized all personal information in the benchmark.

\subsection{\bench-2403}
\label{sec:bench:2403}

Through the above pipeline, we collect and release the first version - \bench-2403. The statistics of \bench-2403 are shown in Table \ref{tab:benchmark_comparison}. We discuss its features as follows.

\begin{table*}[t]
\centering
\caption{The comparison between existing benchmarks and \bench-2403. \texttt{SA} and \texttt{Depend} are the abbreviations of ``standalone'' and ``dependency'', respectively.}
\label{tab:benchmark_comparison}
\resizebox{0.9\linewidth}{!}{
\begin{tabular}{l|cccc|cc|c}
\toprule
\multirow{2}{*}{Benchmark} & \multicolumn{4}{c}{Code Distribution} & \multicolumn{2}{|c|}{Dependency Distribution} & \multirow{2}{*}{Annotation} \\ 
 & \#Repo. & \#Sample & SA & Non-SA & \#Type & \#Avg. &   \\ \midrule
CoNaLa \cite{CoNaLA} & -- & 500 & 100\% & 0\% & 0 & 0 & NL, Code  \\
HumanEval \cite{Codex} & -- & 164 & 100\% & 0\% & 0 & 0 & NL, Code  \\
MBPP \cite{MBPP} & -- & 974 & 100\% & 0\% & 0 & 0 & NL, Code  \\
APPS \cite{APPS} & -- & 5,000 & 100\% & 0\% & 0 & 0 & NL, Code  \\
PandasEval \cite{CERT} & -- & 101 & 100\% & 0\% & 0 & 0 & NL, Code \\
NumpyEval \cite{CERT} & -- & 101 & 100\% & 0\% & 0 & 0 & NL, Code  \\
AixBench \cite{SkCoder} & -- & 175 & 100\% & 0\% & 0 & 0 & NL, Code  \\
ClassEval \cite{ClassEval} & -- & 100 & 100\% & 0\% & 0 & 0 & NL, Code, Depend. Name \\
\midrule
Concode \cite{Concode} & -- & 2,000 & 20\% & 80\% & 1 & 1.23 & NL, Code  \\
CoderEval \cite{CoderEval} & 43 & 230 & 36\% & 64\% & 3 & 1.73 & NL, Code, Depend. Name  \\
DevEval \cite{DevEval} & 117 & 1,874 & 27\% & 73\% & 3 & 3.41 & NL, Code, Depend, Repo \\
\rowcolor[rgb]{ .741,  .843,  .933}
EvoCodeBench-2403 & 25 & 275 & 27\% & 73\% & 3 & 3.46 &\begin{tabular}[c]{@{}c@{}}NL, Code, Depend\\ Repo, Domain\end{tabular} \\ \midrule
500 Real Repositories & 500 & 1M+ & 27\% & 73\% & 3 & 3.22 & -- \\
\bottomrule
\end{tabular}}
\end{table*}

\ding{182} \textbf{Latest repositories to avoid data leakage.} Considering that most code LLM's \cite{StarCoder-2,DeepSeek_Coder} training data is up to September 2023, existing benchmarks might have been leaked. For example, all repositories in CoderEval were created before September 2023.
In contrast, the 25 repositories in \bench-2403 were created between October 2023 and March 2024 and are not included in the training data. The details of 25 repositories is in Appendix \ref{sec:appendix:collection:projects}.

\ding{183} \textbf{Diverse domains.} \bench-2403 covers all programming domains in our taxonomy. The domain distribution of \bench-2403 is shown in Table \ref{tab:domain_distribution}. It provides a broad platform to evaluate and analyze the performance of LLMs across domains. Because that \bench-2403 is our first version, the domain distribution may be unbalanced. In the future, we will collect new samples from the latest repositories and expand the data sizes in different domains.

\ding{184} \textbf{High data quality.} \bench-2403 is collected by a rigorous pipeline and contains high-quality test data. First, as shown in Table \ref{tab:benchmark_comparison}, \bench-2403 aligns with real-world repositories in multiple aspects. 
For example, the code distribution of \bench-2403 is consistent with that of 500 real-world repositories\footnote{We reuse the statistics of 500 real-world repositories reported in related work \cite{DevEval}.}. Second, \bench-2403 provides comprehensive annotations (\eg requirements, reference code, reference dependency, and the original repository). These annotations offer a broad arena to explore repo-level code generation. Third, each sample in \bench-2403 is equipped with an average of 6 test cases rigorously validated through human reviews. In comparison, each sample in a popular benchmark - MBPP \cite{MBPP} has three test cases on average.

\section{Experiments}
\label{sec:experiments}

\subsection{Studied LLMs}
\label{sec:experiments:base_llms}

We select 8 popular LLMs and evaluate them in \bench. They cover general LLMs (\ie gpt-4-turbo-1106 \cite{gpt-4} and gpt-3.5-turbo-1106 \cite{gpt-3.5}) and Code LLMs (\ie StarCoder 2-\{15B, 7B\} \cite{StarCoder-2}, DeepSeek Coder-\{33B, 6.7B\} \cite{DeepSeek_Coder}, and CodeLLaMa-\{13B, 7B\} \cite{CodeLLaMa}). We use official interfaces or implementations to reproduce these LLMs. We run these LLMs on 4 NVIDI A100-40GB GPUs.

\begin{wraptable}{r}{0.55\linewidth}
\vspace{-18mm}
\centering
\caption{The results of data leakage detection.}
\label{tab:CDD_Results}
\resizebox{\linewidth}{!}{
\begin{tabular}{llc}
\toprule
Benchmark & LLMs & Leak Ratio (\%) $\downarrow$ \\ \midrule
HumanEval & gpt-3.5 & \textbf{41.47} \\ \midrule
\multirow{8}{*}{EvoCodeBench-2403} 
 & gpt-4 & 2.18 \\
 & gpt-3.5 & 1.75 \\
 & DeepSeek Coder-33B & 1.88 \\
 & DeepSeek Coder-7B & 1.82 \\
 & StarCoder 2-15B & 1.45 \\
 & StarCoder 2-7B & 1.09 \\
 & CodeLLaMa-13B & 0.82 \\
 & CodeLLaMa-7B & 0.73 \\
 \bottomrule
\end{tabular}}
\vspace{-4mm}
\end{wraptable}

\subsection{Data Leakage Detetion}
\label{sec:experiments:leakage_detection}

As stated in Section \ref{sec:bench}, an advantage of \bench is to alleviate data leakage significantly. To validate this point, we use the latest data leakage detection approach - CDD \cite{CDD} to check \bench-2403. CDD can detect whether LLMs have been trained on specific benchmarks and their variants. The detection results are shown in Table \ref{tab:CDD_Results}. Compared to a mainstream benchmark - HumanEval \cite{Codex}, the leakage rate of \bench-2403 drops significantly to less than 3\%. Besides, since different repositories typically contain similar programs (\eg logging functions), it is almost impossible to achieve a 0\% leakage rate. 

Thus, we think that \bench-2403 is leakage-free and can provide trustworthy evaluations in repo-level code generation.

\subsection{Performance in Repo-level Code Generation}

\noindent \textbf{Experimental Settings.} Repo-level code generation takes a requirement and a repository as inputs. Typically, a repository is very long and surpasses the context windows of existing LLMs. Following previous work \cite{DevEval,CrossCodeEval}, we extract parts of code contexts from the repository as inputs and design the following experimental settings. \textbf{\ding{182} Without context.} We ignore contexts and directly generate the code based on requirements and signatures. \textbf{\ding{183} Local File (Completion).} The local file denotes the code file where the reference code is in. This setting simulates the scenario where developers continue to write code at the end of a file. Besides the requirements and signatures, LLMs can access code contexts above the reference code in the local file.
\textbf{\ding{184} Local File (Infilling).} This setting simulates the scenario where developers infill code in the middle of a file. Besides requirements and signatures, LLMs can see the code contexts above and below the reference code in the local file.

We use Pass@$k$ and Recall@$k$ (see Section \ref{sec:bench:task_metric}) to assess generated programs. In this paper, $k \in [1, 3, 5, 10]$. When $k=1$, we use the greedy search and generate a single program per requirement. When $k>1$, we use the nucleus sampling with a temperature 0.4 and sample 20 programs per requirement. We set the top-$p$ to 0.95 and the max generation length to 500. Because \bench is an evolving benchmark, this paper evaluates LLMs upon \bench-2403. Note that the Pass@$k$ and Recall@$k$ between different versions of \bench are not comparable.

\begin{table*}[t]
\caption{Pass@$k$ and Recall@$k$ of LLMs on \bench-2403. Bold and underlined data indicate top-1 and top-2 results, respectively.}
\label{tab:main_results}
\resizebox{\linewidth}{!}{
\begin{tabular}{lc|cccc|cccc}
\toprule
LLMs & Size & Pass@1 & Pass@3 & Pass@5 & Pass@10 & Recall@1 & Recall@3 & Recall@5 & Recall@10 \\
\midrule
\rowcolor[rgb]{ .741,  .843,  .933}
\multicolumn{10}{c}{Local File (Infilling)} \\
\midrule
gpt-4 & N/A & \textbf{20.73} & \textbf{23.03} & \textbf{24.11} & \textbf{25.34} & 68.24 & 70.63 & 72.05 & 73.52 \\
gpt-3.5 & N/A & 17.82 & 21.78 & 23.06 & 24.46 & 61.94 & 68.13 & 69.69 & 70.85 \\
DeepSeek Coder & 33B & {\ul 19.64} & {\ul 22.78} & {\ul 24.29} & {\ul 26.01} & \textbf{71.46} & \textbf{79.93} & \textbf{82.11} & \textbf{86.25} \\
DeepSeek Coder & 6.7B & 17.82 & 21.02 & 22.40 & 23.97 & {\ul 69.58} & {\ul 74.04} & {\ul 78.00} & {\ul 83.22} \\
StarCoder 2 & 15B & 15.27 & 17.54 & 18.63 & 20.09 & 50.90 & 53.29 & 55.89 & 61.76 \\
StarCoder 2 & 7B & 14.91 & 17.29 & 18.63 & 19.86 & 56.35 & 60.59 & 63.74 & 74.20 \\
\midrule
\rowcolor[rgb]{ .741,  .843,  .933}
\multicolumn{10}{c}{Local File (Completion)} \\
\midrule
gpt-4 & N/A & \textbf{17.45} & \textbf{19.65} & \textbf{20.80} & \textbf{22.41} & 63.49 & 68.67 & 70.00 & 72.07 \\
gpt-3.5 & N/A & {\ul 15.64} & 17.29 & 18.21 & 19.36 & 61.44 & 66.25 & 66.82 & 69.89 \\
DeepSeek Coder & 33B & 14.18 & {\ul 17.57} & 18.66 & 19.95 & {\ul 66.90} & \textbf{72.83} & 74.40 & 80.02 \\
DeepSeek Coder & 6.7B & 13.45 & 17.10 & {\ul 18.81} & {\ul 21.07} & 65.76 & {\ul 72.32} & {\ul 75.61} & 78.45 \\
StarCoder 2 & 15B & 13.82 & 15.44 & 17.84 & 19.59 & \textbf{68.55} & 71.37 & 74.76 & 77.70 \\
StarCoder 2 & 7B & 13.45 & 15.15 & 16.18 & 17.65 & 62.93 & 69.85 & 73.54 & 78.40 \\
CodeLLaMa & 13B & 12.73 & 15.78 & 16.86 & 18.19 & 63.34 & 71.26 & \textbf{76.43} & {\ul 80.11} \\
CodeLLaMa & 7B & 12.73 & 15.33 & 16.00 & 16.93 & 63.33 & 69.79 & 71.91 & 76.50 \\
\midrule
\rowcolor[rgb]{ .741,  .843,  .933}
\multicolumn{10}{c}{Without Context} \\ \midrule
gpt-4 & N/A & \textbf{7.27} & \textbf{10.05} & \textbf{10.70} & {\ul 11.49} & 21.58 & 23.93 & 25.69 & 26.23 \\
gpt-3.5 & N/A & 6.55 & 7.85 & 8.28 & 8.73 & 21.66 & 24.31 & 24.77 & 25.40 \\
DeepSeek Coder & 33B & {\ul 6.91} & {\ul 8.92} & 9.79 & 11.03 & \textbf{27.67} & \textbf{32.73} & {\ul 34.92} & {\ul 37.76} \\
DeepSeek Coder & 6.7B & 5.82 & 8.56 & 9.67 & 11.26 & 25.89 & 32.06 & \textbf{35.59} & \textbf{38.33} \\
StarCoder 2 & 15B & 6.18 & 8.77 & {\ul 9.95} & \textbf{11.53} & 24.03 & 29.86 & 33.62 & 36.91 \\
StarCoder 2 & 7B & 5.82 & 6.72 & 7.43 & 8.62 & {\ul 27.39} & {\ul 32.60} & 34.88 & 36.81 \\
CodeLLaMa & 13B & 5.45 & 7.38 & 8.37 & 9.95 & 25.52 & 31.28 & 33.66 & 36.36 \\
CodeLLaMa & 7B & 5.45 & 6.94 & 7.75 & 9.03 & 26.97 & 31.17 & 34.08 & 36.82 \\
\bottomrule
\end{tabular}}
\end{table*}

\noindent \textbf{Results.} The Pass@$k$ and Recall@$k$ of studied LLMs are shown in Table \ref{tab:main_results}. 

\textbf{Compared to previous benchmarks, these LLMs' performance in \bench-2403 drops dramatically.} For example, the highest Pass@1 scores of gpt-4 on the latest repo-level benchmark \cite{DevEval} is 53.04. In contrast, gpt-4 only achieves 20.73 on Pass@1 upon \bench-2403. The decreases demonstrate that \bench is more challenging, and previous benchmarks might have been leaked.

\begin{figure}[t]
\centering
\includegraphics[width=\linewidth]{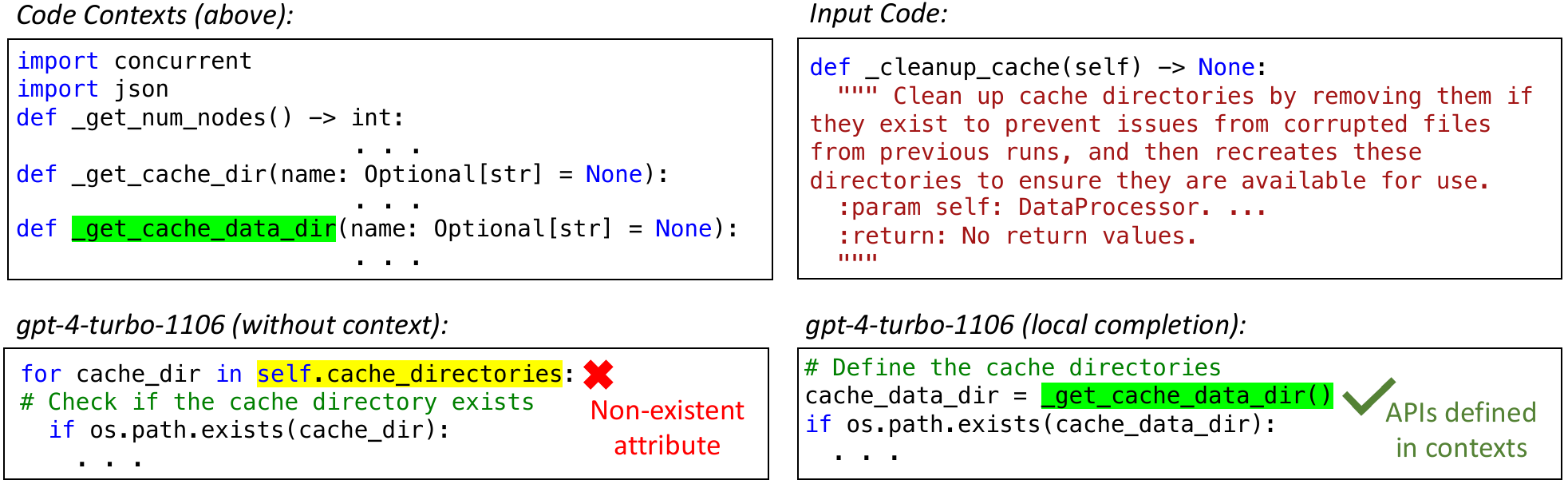}
\caption{A uniquely successful case in Local File (Completion) setting.}
\label{fig:success_case}
\end{figure}

\textbf{LLMs benefit from more code contexts in repo-level code generation.} As shown in Table \ref{tab:main_results}, after introducing the contexts, the Pass@$k$ and Recall@$k$ of LLMs obviously increase. For example, the Pass@1 of gpt-4 is improved by 104\% and 152\% in two settings, respectively. Similar phenomena occur in previous studies \cite{DevEval}. We attribute the improvements to the domain knowledge contained in contexts. Figure \ref{fig:success_case} shows a uniquely successful case in the Local File (Completion) setting. Without context, gpt-4 fabricated a non-existent field as cache directories, generating the incorrect code. In fact, two functions for returning the cache directories are available in the local file. After introducing the local file, gpt-4 successfully invokes relevant functions and generates the correct code. 

\begin{wraptable}{r}{0.5\linewidth}
\vspace{-7mm}
\centering
\caption{Performance of RAG.}
\label{tab:RAG}
\resizebox{\linewidth}{!}{
\begin{tabular}{llcc}
\toprule
LLMs & Setting & Pass@1 & Recall@1 \\
\midrule
\multirow{2}{*}{gpt-4} & Without Context & 8.31 & 21.08 \\
 & Similar Functions & 12.29 & 45.14 \\ \midrule
\multirow{2}{*}{gpt-3.5} & Without Context & 6.64 & 21.16 \\
 & Similar Functions & 11.62 & 41.93 \\
 \bottomrule
\end{tabular}}
\end{wraptable}

\textbf{Retrieval-Augmented Generation (RAG).} RAG enhances generative models with retrieved information and has achieved promising results in code generation \cite{SkCoder,AceCoder}. We apply RAG to repo-level code generation and consider the current repository a retrieval corpus. Because most programs in repositories are not equipped with documentation, we retrieve top-$k$ (\ie $k=5$ in this paper) functions with similar names to the target function. Specifically, we split names into tokens based on underscore or camelcase formatting and then match the tokens of names. Finally, we use similar functions as contexts in prompts. The results are shown in Table \ref{tab:RAG}. The performance of both LLMs is improved after introducing similar functions. We attribute the improvements to relevant algorithms and dependencies in similar functions. It inspired researchers to explore more advanced RAG techniques to improve repo-level code generation.

\noindent \textbf{Error Analyses.} The Pass@$k$ of existing LLMs in repo-level code generation is still low. To determine LLMs' shortcomings, we manually analyze 50 error cases of gpt-4 in the Local File (Infilling) setting. We found the most cases (29 cases) are caused by implementation logic errors. 20 cases failed since the necessary contexts were missing, \eg APIs defined in other files. Besides, one case failed because of the vague requirement. It shows that existing LLMs' reasoning and coding abilities need to be improved. Meanwhile, how to utilize more contexts is necessary to explore.

\subsection{Performance in Different Domains}
\label{sec:experiments:results_domain}

\begin{table}[t]
\centering
\caption{The Pass@1 of studied LLMs in different domains upon \bench-2403.}
\label{tab:domain_results}
\resizebox{0.85\linewidth}{!}{
\begin{tabular}{lcccccccc}
\toprule
\multirow{2}{*}{Domain} & \multirow{2}{*}{gpt-4} & \multirow{2}{*}{gpt-3.5} & \multicolumn{2}{c}{DeepSeek Coder} & \multicolumn{2}{c}{StarCoder 2} & \multicolumn{2}{c}{CodeLLaMa} \\ 
 &  &  & 33B & 6.7B & 15B & 7B & 13B & 7B \\ \midrule
Database & \textbf{38.89} & \textbf{38.89} & 33.33 & 33.33 & \textbf{38.89} & 33.33 & 33.33 & 33.33 \\
System & \textbf{35.29} & \textbf{35.29} & \textbf{35.29} & \textbf{35.29} & 29.41 & 29.41 & 23.53 & \textbf{35.29} \\
Software Development & \textbf{12.00} & \textbf{12.00} & 8.00 & \textbf{12.00} & 10.00 & 6.00 & 8.00 & 8.00 \\
Internet & 20.00 & \textbf{26.67} & \textbf{26.67} & \textbf{26.67} & 20.00 & \textbf{26.67} & \textbf{26.67} & \textbf{26.67} \\
Scientific Engineering & \textbf{11.67} & 10.00 & 10.00 & 6.67 & 8.33 & 9.17 & 7.50 & 8.33 \\
Multimedia & \textbf{25.00} & 18.75 & 15.63 & 15.63 & 18.75 & 18.75 & 18.75 & 12.50 \\
Text Processing & \textbf{16.67} & 0.00 & 0.00 & 0.00 & 0.00 & 0.00 & 0.00 & 0.00 \\ \midrule
All Domains & 17.45 & 15.64 & 14.18 & 13.45 & 13.82 & 13.45 & 12.73 & 12.73 \\
\bottomrule
\end{tabular}}
\end{table}

We divide \bench into multiple subsets according to the domain labels and then calculate the Pass@1 of LLMs in different domains. The results are shown in Table \ref{tab:domain_results}. We ignore three domains with less than 10 samples and leave them for future work.

\textbf{\bench shows superior LLMs in specific domains.} The Pass@1 scores in overall benchmarks demonstrate the comprehensive coding abilities of LLMs. 
Because developers typically focus on specific programming domains, they are more concerned about the performance of LLMs in specific domains. Imagine we are developers focused on internet-related programming tasks. Based on the overall Pass@1, we would think StarCoder 2-7B is stronger than DeepSeek Coder-6.7B, \ie 13.82 $>$ 13.45. However, according to Table \ref{tab:domain_results}, DeepSeek Coder-6.7B performs better than StarCoder 2-7B in the Internet domain. This result can help us to select more suitable models.

\begin{table}[t]
\centering
\caption{The Domain-Specific Improvements (\%) of LLMs in different domains. The comfort domains and strange domains are marked in bleu and red, respectively.}
\label{tab:DSI_Results}
\resizebox{0.85\linewidth}{!}{
\begin{tabular}{lcccccccc}
\toprule
 &  &  & \multicolumn{2}{c}{DeepSeek Coder} & \multicolumn{2}{c}{StarCoder 2} & \multicolumn{2}{c}{CodeLLaMa} \\
\multirow{-2}{*}{Domain} & \multirow{-2}{*}{gpt-4} & \multirow{-2}{*}{gpt-3.5} & 33B & 6.7B & 15B & 7B & 13B & 7B \\ \midrule
Database & \cellcolor[rgb]{ .741,  .843,  .933}10.21 & \cellcolor[rgb]{ .741,  .843,  .933}10.21 & -7.14 & -7.14 & -7.14 & \cellcolor[rgb]{ .741,  .843,  .933}10.21 & -7.15 & -7.15 \\
System & 9.51 & 9.52 & 9.51 & 9.51 & \cellcolor[rgb]{ 1,  .7,  .7}-11.42 & \cellcolor[rgb]{ 1,  .7,  .7}-11.42 & \cellcolor[rgb]{ 1,  .7,  .7}-42.84 & 9.52 \\
Software Development & \cellcolor[rgb]{ .741,  .843,  .933}23.81 & \cellcolor[rgb]{ .741,  .843,  .933}23.81 & \cellcolor[rgb]{ 1,  .7,  .7}-21.43 & \cellcolor[rgb]{ .741,  .843,  .933}23.81 & \cellcolor[rgb]{ 1,  .7,  .7}-66.67 & 5.71 & \cellcolor[rgb]{ 1,  .7,  .7}-21.43 & \cellcolor[rgb]{ 1,  .7,  .7}-21.43 \\
Internet & \cellcolor[rgb]{ 1,  .7,  .7}-28.59 & 7.15 & 7.15 & 7.15 & 7.15 & \cellcolor[rgb]{ 1,  .7,  .7}-28.59 & 7.15 & 7.15 \\
Scientific Engineering & \cellcolor[rgb]{ .741,  .843,  .933}26.55 & \cellcolor[rgb]{ .741,  .843,  .933}11.90 & \cellcolor[rgb]{ .741,  .843,  .933}11.90 & \cellcolor[rgb]{ 1,  .7,  .7}-39.22 & 2.63 & -8.63 & \cellcolor[rgb]{ 1,  .7,  .7}-22.23 & -8.63 \\
Multimedia & \cellcolor[rgb]{ .741,  .843,  .933}32.14 & 4.75 & \cellcolor[rgb]{ 1,  .7,  .7}-17.11 & \cellcolor[rgb]{ 1,  .7,  .7}-17.11 & 4.75 & 4.75 & 4.75 & \cellcolor[rgb]{ 1,  .7,  .7}-50.01 \\
Text Processing & \cellcolor[rgb]{ .741,  .843,  .933}100.00 & \cellcolor[rgb]{ 1,  .7,  .7}-100 & \cellcolor[rgb]{ 1,  .7,  .7}-100 & \cellcolor[rgb]{ 1,  .7,  .7}-100 & \cellcolor[rgb]{ 1,  .7,  .7}-100 & \cellcolor[rgb]{ 1,  .7,  .7}-100 & \cellcolor[rgb]{ 1,  .7,  .7}-100 & \cellcolor[rgb]{ 1,  .7,  .7}-100 \\
\bottomrule
\end{tabular}}
\vspace{-4mm}
\end{table}

\textbf{\bench uncovers the comfort domains and strange domains of specific LLMs.} For ease of observation, we compute the Domain-Specific Improvement (DSI) of LLMs in different domains. The DSI refers to the average relative improvement of Pass@1 of an LLM in a domain compared to other LLMs. Suppose we evaluate $N$ LLMs on a specific domain, and their Pass@1 scores are represented as $\mathbf{P}$. Then, the DSI (\%) of $i$-th LLM in this domain is computed as:
\begin{equation}
DSI_i = \frac{1}{N-1}\sum_j \frac{P_i-P_j}{P_i} * 100 \quad (i \neq j)
\end{equation}

The larger the DSI, the better an LLM is at that domain. If an LLM's DSI in a domain exceeds a threshold $\mathbf{T}$, we consider it a comfort domain. If an LLM's DSI in a domain is less than $\mathbf{-T}$, it is considered a strange domain. $\mathbf{T}$ is a hyper-parameter and is set to 10\% in this paper. Practitioners can further tune this parameter.

Table \ref{tab:DSI_Results} shows the DSIs of studied LLMs across domains. The comfort domains and strange domains are marked in bleu and red, respectively. We can see that gpt-4 has the most comfort domains. Especially in the Text Processing domain, among all LLMs, only gpt-4 successfully solves some programming tasks. However, gpt-4 performs worse than others in the Internet domain. Besides, we discover that StarCoder 2-15B unexpectedly performs well in the Database domain and even is comparable to gpt-4. The potential reason for comfort and strange domains is that the pre-training data mix of LLMs is different. 
For example, gpt-4's training data contains fewer repositories in the Internet domain, resulting in weak performance.
These findings can help model trainers analyze the shortcomings of existing LLMs and build more powerful code LLMs.

\section{Discussion}
\label{sec:discussion}

\noindent \textbf{Evaluation of auto-generated annotations.} We leverage an LLM (\ie gpt-4 in this paper) to annotate natural language requirements and domain labels for functions. To assess the quality of auto-generated annotations, We hire five developers to write requirements and domain labels for \bench-2403. Then, we hire another five developers to compare annotations from gpt-4 and developers. All of these developers have at least 3 years of Python development experience.
All developers are paid according to the relevant policies\footnote{\url{https://www.worker.gov/}} (\ie \$7.5 per hour). The details of human evaluation are in Appendix \ref{sec:appendix:experiment:human_eval}.


\begin{wraptable}{r}{0.6\linewidth}
\vspace{-12mm}
\caption{Human evaluation of auto-generated annotations.}
\label{tab:human_evaluation}
\resizebox{\linewidth}{!}{
\begin{tabular}{lcccc}
\toprule
\multirow{2}{*}{Annotator} & \multicolumn{2}{c}{Win / Tie / Lose} & \multirow{2}{*}{Cost (Time)} & \multirow{2}{*}{Cost (Money)} \\
 & Requirement & Domain &  &  \\ \midrule
gpt-4-turbo-1106 & 30 / 236 / 9 & 3 / 268 / 4 & 1h9m & \$3.11 \\
Human & 9 / 236 / 30 & 4 / 268 / 3 & 23h & \$172.5 \\
\bottomrule
\end{tabular}}
\vspace{-4mm}
\end{wraptable}

The evaluation results are shown in Table \ref{tab:human_evaluation}. The Cohen's Kappa coefficient between all evaluators is 0.9. The \textit{Tie} means both requirements are satisfying.
We can see that gpt-4 produces high-quality annotations comparable to human-written annotations in most cases (\eg requirement: 96.7\% = (30+236)/275, domain labels: 98.5\% = (3+268)/275).
We also inspect failed cases of gpt-4 and summarize two main reasons. First, gpt-4 may miss some details (\eg hyper-parameters) that are necessary for requirements. Second, gpt-4 may be mistaken by specific APIs and output inaccurate domain labels. 
In the future, we will explore new techniques to solve this problem, \eg controllable text generation \cite{control_gen}.
Besides, gpt-4 shows advantages in costs. As shown in Table \ref{tab:human_evaluation}, gpt-4 costs less time and money to annotate requirements. Thus, it is a feasible and efficient approach for us to use gpt-4 to annotate requirements for \bench.

\noindent \textbf{Limitations.} There are two main limitations in \bench. First, \bench is a monolingual (\ie Python) benchmark and ignores other programming languages (\eg Java, C++). Because building repo-level benchmarks faces many language-specific challenges (\eg how to install execution environments, how to run test cases), we chose to start with Python, a mainstream programming language in existing benchmarks \cite{CoNaLA,Codex,ClassEval,CoderEval}. We plan to support other programming languages in the future gradually. Second, the size of \bench is currently smaller than some existing benchmarks. The reason is that \bench-2403 only collects samples from recent repositories (\ie Oct. 2023 - Mar. 2024). In the future, we will continue to collect new samples from the latest repositories and expand the scale of \bench.

\section{Related Work}
\label{sec:related_work}

\noindent \textbf{Code Generation Benchmarks.} 
Nowadays, prevalent code generation benchmarks can be divided into two groups: \ding{182} Snippet-level benchmarks \cite{Codex,MBPP,APPS,ClassEval}. They comprise human-written or competitive programming problems, which ask LLMs to generate standalone code snippets. \ding{183} Repo-level benchmarks \cite{CoderEval,DevEval}. They ask LLMs to generate new programs based on requirements and contexts from current repositories. Compared to snippet-level benchmarks, repo-level benchmarks are more challenging and closer to real-world software development scenarios. 

This paper proposes a new benchmark - \bench, to alleviate two limitations of previous benchmarks (\ie data leakage and lack of domain-specific evaluations). We notice that some recent benchmarks focus on similar limitations. We further clarify the differences between \bench and existing benchmarks.

\noindent \textbf{Data leakage.} LiveCodeBench \cite{LiveCodeBench} collects the latest competitive programming problems. EvoEval \cite{EvoEval} leverages LLMs to mutate HumanEval and obtain new benchmarks. They are both snippet-level benchmarks, while \bench is a more practical repo-level benchmark. The collection pipelines in LiveCodeBench and EvoEval can not be applied to repo-level benchmarks, which involve many new challenges (\eg repository selection, test construction, and requirement annotation). We fill this knowledge gap by building a new collection pipeline and release \bench.

\noindent \textbf{Domain-specific evaluations.} Existing benchmarks typically fall into narrow domains (\eg PandasEval \cite{CERT}) or lack domain labels (\eg DevEval \cite{DevEval}). ClassEval \cite{ClassEval} contains 100 standalone programming tasks from seven domains. However, these domains are manually designed based on human experiences and may ignore important domains (\eg Internet and Multimedia). Besides, ClassEval ignores repo-level benchmarks and may be leaked in the future. In contrast, we consider the statistics of a mainstream open-source community and identify the top 10 popular programming domains. Besides, \bench is free of data leakage and continually expands domains.

\section{Conclusion and Future Work}
\label{sec:conclusion}

We introduce \bench, an evolving code generation benchmark. \bench 
is designed to address two limitations (\ie data leakage and lack of domain-specific evaluations). \bench is an evolving benchmark and will be dynamically updated every period (\eg six months), to avoid data leakage. This paper releases the first version - \bench-2403, which contains 275 samples.
Besides, we design a programming domain taxonomy consisting of ten popular domains and annotate samples with domain labels. We conduct extensive experiments on \bench and reveal the actual abilities of LLMs in real-world repositories. We also evaluate LLMs in different domains and discover their comfort and strange domains. These insights can help practitioners evaluate LLMs comprehensively.

In the future, we will continuously release new versions of \bench and extend \bench into other programming languages (\eg Java and C++).

\section*{Acknowledgements}

This research is supported by the National Natural Science Foundation of China (Nos. 62192731, 62152730), the National Key R\&D Program under Grant No. 2023YFB4503801, the National Natural Science Foundation of China (Nos. 62072007, 62192733, 61832009, 62192730), and the Major Program (JD) of Hubei Province (No.2023BAA024). Ge Li and Yongbing Li are the corresponding authors.

\bibliography{custom}

\begin{thebibliography}{10}

\bibitem{MBPP}
Jacob Austin, Augustus Odena, Maxwell~I. Nye, Maarten Bosma, Henryk Michalewski, David Dohan, Ellen Jiang, Carrie~J. Cai, Michael Terry, Quoc~V. Le, and Charles Sutton.
\newblock Program synthesis with large language models.
\newblock {\em CoRR}, abs/2108.07732, 2021.

\bibitem{DBLP:journals/corr/abs-2303-12712}
S{\'{e}}bastien Bubeck, Varun Chandrasekaran, Ronen Eldan, Johannes Gehrke, Eric Horvitz, Ece Kamar, Peter Lee, Yin~Tat Lee, Yuanzhi Li, Scott~M. Lundberg, Harsha Nori, Hamid Palangi, Marco~T{\'{u}}lio Ribeiro, and Yi~Zhang.
\newblock Sparks of artificial general intelligence: Early experiments with {GPT-4}.
\newblock {\em CoRR}, abs/2303.12712, 2023.

\bibitem{Codex}
Mark Chen, Jerry Tworek, Heewoo Jun, Qiming Yuan, Henrique~Pond{\'{e}} de~Oliveira~Pinto, Jared Kaplan, Harrison Edwards, Yuri Burda, Nicholas Joseph, Greg Brockman, Alex Ray, Raul Puri, Gretchen Krueger, Michael Petrov, Heidy Khlaaf, Girish Sastry, Pamela Mishkin, Brooke Chan, Scott Gray, Nick Ryder, Mikhail Pavlov, Alethea Power, Lukasz Kaiser, Mohammad Bavarian, Clemens Winter, Philippe Tillet, Felipe~Petroski Such, Dave Cummings, Matthias Plappert, Fotios Chantzis, Elizabeth Barnes, Ariel Herbert{-}Voss, William~Hebgen Guss, Alex Nichol, Alex Paino, Nikolas Tezak, Jie Tang, Igor Babuschkin, Suchir Balaji, Shantanu Jain, William Saunders, Christopher Hesse, Andrew~N. Carr, Jan Leike, Joshua Achiam, Vedant Misra, Evan Morikawa, Alec Radford, Matthew Knight, Miles Brundage, Mira Murati, Katie Mayer, Peter Welinder, Bob McGrew, Dario Amodei, Sam McCandlish, Ilya Sutskever, and Wojciech Zaremba.
\newblock Evaluating large language models trained on code.
\newblock {\em CoRR}, 2021.

\bibitem{control_gen}
Jasper Dekoninck, Marc Fischer, Luca Beurer{-}Kellner, and Martin~T. Vechev.
\newblock Controlled text generation via language model arithmetic.
\newblock {\em CoRR}, abs/2311.14479, 2023.

\bibitem{CrossCodeEval}
Yangruibo Ding, Zijian Wang, Wasi~Uddin Ahmad, Hantian Ding, Ming Tan, Nihal Jain, Murali~Krishna Ramanathan, Ramesh Nallapati, Parminder Bhatia, Dan Roth, and Bing Xiang.
\newblock Crosscodeeval: {A} diverse and multilingual benchmark for cross-file code completion.
\newblock In Alice Oh, Tristan Naumann, Amir Globerson, Kate Saenko, Moritz Hardt, and Sergey Levine, editors, {\em Advances in Neural Information Processing Systems 36: Annual Conference on Neural Information Processing Systems 2023, NeurIPS 2023, New Orleans, LA, USA, December 10 - 16, 2023}, 2023.

\bibitem{CDD}
Yihong Dong, Xue Jiang, Huanyu Liu, Zhi Jin, Bin Gu, Mengfei Yang, and Ge~Li.
\newblock Generalization or memorization: Data contamination and trustworthy evaluation for large language models.
\newblock In Lun-Wei Ku, Andre Martins, and Vivek Srikumar, editors, {\em Findings of the Association for Computational Linguistics ACL 2024}, pages 12039--12050, Bangkok, Thailand and virtual meeting, August 2024. Association for Computational Linguistics.

\bibitem{ClassEval}
Xueying Du, Mingwei Liu, Kaixin Wang, Hanlin Wang, Junwei Liu, Yixuan Chen, Jiayi Feng, Chaofeng Sha, Xin Peng, and Yiling Lou.
\newblock Evaluating large language models in class-level code generation.
\newblock In {\em Proceedings of the 46th {IEEE/ACM} International Conference on Software Engineering, {ICSE} 2024, Lisbon, Portugal, April 14-20, 2024}, pages 81:1--81:13. {ACM}, 2024.

\bibitem{red-teaming}
Deep Ganguli, Liane Lovitt, Jackson Kernion, Amanda Askell, Yuntao Bai, Saurav Kadavath, Ben Mann, Ethan Perez, Nicholas Schiefer, Kamal Ndousse, Andy Jones, Sam Bowman, Anna Chen, Tom Conerly, Nova DasSarma, Dawn Drain, Nelson Elhage, Sheer~El Showk, Stanislav Fort, Zac Hatfield{-}Dodds, Tom Henighan, Danny Hernandez, Tristan Hume, Josh Jacobson, Scott Johnston, Shauna Kravec, Catherine Olsson, Sam Ringer, Eli Tran{-}Johnson, Dario Amodei, Tom Brown, Nicholas Joseph, Sam McCandlish, Chris Olah, Jared Kaplan, and Jack Clark.
\newblock Red teaming language models to reduce harms: Methods, scaling behaviors, and lessons learned.
\newblock {\em CoRR}, abs/2209.07858, 2022.

\bibitem{DataSheet}
Timnit Gebru, Jamie Morgenstern, Briana Vecchione, Jennifer~Wortman Vaughan, Hanna~M. Wallach, Hal~Daum{\'{e}} III, and Kate Crawford.
\newblock Datasheets for datasets.
\newblock {\em Commun. {ACM}}, 64(12):86--92, 2021.

\bibitem{DeepSeek_Coder}
Daya Guo, Qihao Zhu, Dejian Yang, Zhenda Xie, Kai Dong, Wentao Zhang, Guanting Chen, Xiao Bi, Y.~Wu, Y.~K. Li, Fuli Luo, Yingfei Xiong, and Wenfeng Liang.
\newblock Deepseek-coder: When the large language model meets programming - the rise of code intelligence.
\newblock {\em CoRR}, abs/2401.14196, 2024.

\bibitem{APPS}
Dan Hendrycks, Steven Basart, Saurav Kadavath, Mantas Mazeika, Akul Arora, Ethan Guo, Collin Burns, Samir Puranik, Horace He, Dawn Song, and Jacob Steinhardt.
\newblock Measuring coding challenge competence with {APPS}.
\newblock In Joaquin Vanschoren and Sai{-}Kit Yeung, editors, {\em Proceedings of the Neural Information Processing Systems Track on Datasets and Benchmarks 1, NeurIPS Datasets and Benchmarks 2021, December 2021, virtual}, 2021.

\bibitem{Concode}
Srinivasan Iyer, Ioannis Konstas, Alvin Cheung, and Luke Zettlemoyer.
\newblock Mapping language to code in programmatic context.
\newblock In Ellen Riloff, David Chiang, Julia Hockenmaier, and Jun'ichi Tsujii, editors, {\em Proceedings of the 2018 Conference on Empirical Methods in Natural Language Processing, Brussels, Belgium, October 31 - November 4, 2018}, pages 1643--1652. Association for Computational Linguistics, 2018.

\bibitem{LiveCodeBench}
Naman Jain, King Han, Alex Gu, Wen{-}Ding Li, Fanjia Yan, Tianjun Zhang, Sida Wang, Armando Solar{-}Lezama, Koushik Sen, and Ion Stoica.
\newblock Livecodebench: Holistic and contamination free evaluation of large language models for code.
\newblock {\em CoRR}, abs/2403.07974, 2024.

\bibitem{aiXcoder-7B}
Siyuan Jiang, Jia Li, He~Zong, Huanyu Liu, Hao Zhu, Shukai Hu, Erlu Li, Jiazheng Ding, Yu~Han, Wei Ning, et~al.
\newblock aixcoder-7b: A lightweight and effective large language model for code completion.
\newblock {\em arXiv preprint arXiv:2410.13187}, 2024.

\bibitem{SCoT}
Jia Li, Ge~Li, Yongmin Li, and Zhi Jin.
\newblock Structured chain-of-thought prompting for code generation.
\newblock {\em ACM Trans. Softw. Eng. Methodol.}, August 2024.
\newblock Just Accepted.

\bibitem{DevEval}
Jia Li, Ge~Li, Yunfei Zhao, Yongmin Li, Huanyu Liu, Hao Zhu, Lecheng Wang, Kaibo Liu, Zheng Fang, Lanshen Wang, Jiazheng Ding, Xuanming Zhang, Yuqi Zhu, Yihong Dong, Zhi Jin, Binhua Li, Fei Huang, and Yongbin Li.
\newblock Deveval: A manually-annotated code generation benchmark aligned with real-world code repositories.
\newblock In {\em Proceedings of the 62nd Annual Meeting of the Association for Computational Linguistics}, Bangkok, Thailand, 2024. Association for Computational Linguistics.

\bibitem{SkCoder}
Jia Li, Yongmin Li, Ge~Li, Zhi Jin, Yiyang Hao, and Xing Hu.
\newblock Skcoder: {A} sketch-based approach for automatic code generation.
\newblock In {\em 45th {IEEE/ACM} International Conference on Software Engineering, {ICSE} 2023, Melbourne, Australia, May 14-20, 2023}, pages 2124--2135. {IEEE}, 2023.

\bibitem{AceCoder}
Jia Li, Yunfei Zhao, Yongmin Li, Ge~Li, and Zhi Jin.
\newblock Acecoder: An effective prompting technique specialized in code generation.
\newblock {\em ACM Trans. Softw. Eng. Methodol.}, July 2024.
\newblock Just Accepted.

\bibitem{StarCoder-2}
Anton Lozhkov, Raymond Li, Loubna~Ben Allal, Federico Cassano, Joel Lamy-Poirier, Nouamane Tazi, Ao~Tang, Dmytro Pykhtar, Jiawei Liu, Yuxiang Wei, et~al.
\newblock Starcoder 2 and the stack v2: The next generation.
\newblock {\em arXiv preprint arXiv:2402.19173}, 2024.

\bibitem{gpt-3.5}
OpenAI.
\newblock gpt-3.5-turbo.
\newblock \url{https://platform.openai.com/docs/models/gpt-3-5}, 2023.

\bibitem{gpt-4}
OpenAI.
\newblock {GPT-4} technical report.
\newblock {\em CoRR}, abs/2303.08774, 2023.

\bibitem{Pip}
Pip.
\newblock Pip.
\newblock \url{https://pypi.org/project/pip}, 2024.

\bibitem{Pyan}
Pyan.
\newblock Pyan.
\newblock \url{https://github.com/davidfraser/pyan}, 2023.

\bibitem{PyPI}
PyPI.
\newblock Pypi.
\newblock \url{https://pypi.org/}.

\bibitem{Pytest}
Pytest.
\newblock Pytest.
\newblock \url{https://docs.pytest.org/en/8.0.x/}, 2024.

\bibitem{CodeLLaMa}
Baptiste Rozi{\`{e}}re, Jonas Gehring, Fabian Gloeckle, Sten Sootla, Itai Gat, Xiaoqing~Ellen Tan, Yossi Adi, Jingyu Liu, Tal Remez, J{\'{e}}r{\'{e}}my Rapin, Artyom Kozhevnikov, Ivan Evtimov, Joanna Bitton, Manish Bhatt, Cristian Canton{-}Ferrer, Aaron Grattafiori, Wenhan Xiong, Alexandre D{\'{e}}fossez, Jade Copet, Faisal Azhar, Hugo Touvron, Louis Martin, Nicolas Usunier, Thomas Scialom, and Gabriel Synnaeve.
\newblock Code llama: Open foundation models for code.
\newblock {\em CoRR}, abs/2308.12950, 2023.

\bibitem{EvoEval}
Chunqiu~Steven Xia, Yinlin Deng, and Lingming Zhang.
\newblock Top leaderboard ranking = top coding proficiency, always? evoeval: Evolving coding benchmarks via {LLM}.
\newblock {\em CoRR}, abs/2403.19114, 2024.

\bibitem{CoNaLA}
Pengcheng Yin, Bowen Deng, Edgar Chen, Bogdan Vasilescu, and Graham Neubig.
\newblock Learning to mine aligned code and natural language pairs from stack overflow.
\newblock In Andy Zaidman, Yasutaka Kamei, and Emily Hill, editors, {\em Proceedings of the 15th International Conference on Mining Software Repositories, {MSR} 2018, Gothenburg, Sweden, May 28-29, 2018}, pages 476--486. {ACM}, 2018.

\bibitem{CoderEval}
Hao Yu, Bo~Shen, Dezhi Ran, Jiaxin Zhang, Qi~Zhang, Yuchi Ma, Guangtai Liang, Ying Li, Qianxiang Wang, and Tao Xie.
\newblock Codereval: {A} benchmark of pragmatic code generation with generative pre-trained models.
\newblock In {\em Proceedings of the 46th {IEEE/ACM} International Conference on Software Engineering, {ICSE} 2024, Lisbon, Portugal, April 14-20, 2024}, pages 37:1--37:12. {ACM}, 2024.

\bibitem{CERT}
Daoguang Zan, Bei Chen, Dejian Yang, Zeqi Lin, Minsu Kim, Bei Guan, Yongji Wang, Weizhu Chen, and Jian{-}Guang Lou.
\newblock {CERT:} continual pre-training on sketches for library-oriented code generation.
\newblock In Luc~De Raedt, editor, {\em Proceedings of the Thirty-First International Joint Conference on Artificial Intelligence, {IJCAI} 2022, Vienna, Austria, 23-29 July 2022}, pages 2369--2375. ijcai.org, 2022.

\end{thebibliography}
\bibliographystyle{plain}

\section*{Checklist}

\begin{enumerate}

\item For all authors...
\begin{enumerate}
  \item Do the main claims made in the abstract and introduction accurately reflect the paper's contributions and scope?
    \answerYes{}
  \item Did you describe the limitations of your work?
    \answerYes{See Section~\ref{sec:discussion}.}
  \item Did you discuss any potential negative societal impacts of your work?
    \answerYes{See Section~\ref{sec:discussion}.}
  \item Have you read the ethics review guidelines and ensured that your paper conforms to them?
    \answerYes{}
\end{enumerate}

\item If you are including theoretical results...
\begin{enumerate}
  \item Did you state the full set of assumptions of all theoretical results?
    \answerNA{}
	\item Did you include complete proofs of all theoretical results?
    \answerNA{}
\end{enumerate}

\item If you ran experiments (e.g. for benchmarks)...
\begin{enumerate}
  \item Did you include the code, data, and instructions needed to reproduce the main experimental results (either in the supplemental material or as a URL)?
    \answerYes{See Appendix \ref{sec:appendix:links} and Appendix \ref{sec:appendix:experiment} in the supplemental material.}
  \item Did you specify all the training details (e.g., data splits, hyperparameters, how they were chosen)?
    \answerYes{See Section \ref{sec:experiments}.}
	\item Did you report error bars (e.g., with respect to the random seed after running experiments multiple times)?
    \answerNA{}
	\item Did you include the total amount of compute and the type of resources used (e.g., type of GPUs, internal cluster, or cloud provider)?
    \answerYes{See Section \ref{sec:experiments:base_llms}.}
\end{enumerate}

\item If you are using existing assets (e.g., code, data, models) or curating/releasing new assets...
\begin{enumerate}
  \item If your work uses existing assets, did you cite the creators?
    \answerNA{}
  \item Did you mention the license of the assets?
    \answerYes{We describe the licenses of our collected benchmark in the supplemental material (Appendix \ref{sec:appendix:links}).}
  \item Did you include any new assets either in the supplemental material or as a URL?
    \answerYes{We provide a URL to our collected benchmark in the supplemental material (Appendix \ref{sec:appendix:links}).}
  \item Did you discuss whether and how consent was obtained from people whose data you're using/curating?
    \answerYes{We only consider code repositories with open-source licenses. More details can be found in Section \ref{sec:bench:collection_pipeline}.}
  \item Did you discuss whether the data you are using/curating contains personally identifiable information or offensive content?
    \answerYes{We exclude malicious code repositories. ore details can be found in Section \ref{sec:bench:collection_pipeline}.}
\end{enumerate}

\item If you used crowdsourcing or conducted research with human subjects...
\begin{enumerate}
  \item Did you include the full text of instructions given to participants and screenshots, if applicable?
    \answerYes{We show instructions given to participants. Please see Appendix \ref{sec:appendix:experiment:human_eval} in the supplemental material.}
  \item Did you describe any potential participant risks, with links to Institutional Review Board (IRB) approvals, if applicable?
    \answerNA{}
  \item Did you include the estimated hourly wage paid to participants and the total amount spent on participant compensation?
    \answerYes{All developers are paid according to the relevant policies (i.e., \$7.5 per hour). More details can be found in Line 255.}
\end{enumerate}

\end{enumerate}


\newpage
\appendix

\section*{Appendix}

\section*{Table of Contents}

\begin{itemize}
    \item Appendix~\ref{sec:appendix:links}: Details of hosting, licensing, and maintenance
    \item Appendix~\ref{sec:appendix:statement}: Author responsibility statement
    \item Appendix~\ref{sec:appendix:datasheet}: Dataset datasheet
    \item Appendix \ref{sec:appendix:pipeline}: The details of our collection pipeline
    \item Appendix \ref{sec:appendix:experiment}: The details of experiments
\end{itemize}

\section{Hosting, Licensing, and Maintenance}
\label{sec:appendix:links}
Our \bench and experimental results (\eg code, prompts, and models' predictions) are available on the following platforms.
\begin{itemize}
    \item GitHub: \url{https://github.com/seketeam/EvoCodeBench}
    \item HuggingFace: \url{https://huggingface.co/datasets/LJ0815/EvoCodeBench}
    \item Croissant metadata: \url{https://github.com/seketeam/EvoCodeBench/blob/main/croissant_metadata.json}
\end{itemize}

\bench is available for download under a CC-4.0 license, and our code is available under a BSD 3-Clause license. We ensure the long-term availability and maintenance of the data by hosting it on the GitHub\footnote{\url{https://github.com/}} platform.

\section{Author Responsibility Statement}
\label{sec:appendix:statement}
The authors confirm that they bear all responsibility in case of any rights violation during the data collection or other work and will take appropriate action when needed, \eg to remove data with such issues.
The authors also confirm the licenses provided with the data and code associated with this work.

\section{Datasheet for \bench}
\label{sec:appendix:datasheet}
Questions from Datasheet for Datasets (v8)~\cite{DataSheet}.

\subsection{Motivation}

\textbf{Q: For what purpose was the dataset created?} 

Large Language Models (LLMs) have shown impressive abilities in code generation. This dataset was created to evaluate LLMs in code generation. It specifically fills in two knowledge gaps in previous benchmarks, \ie data leakage and lack of domain-specific evaluations. The former hurts the fairness of benchmarks, and the latter hinders practitioners from selecting superior LLMs for specific domains.

\textbf{Q: Who created the dataset (e.g., which team, research group) and on behalf of which entity (e.g., company, institution, organization)?}

This dataset's authors are from the School of Computer Science at Peking University and the Conversational AI team at Alibaba DAMO Academy.

\textbf{Q: Who funded the creation of the dataset?}

This dataset is funded by Peking University and the Alibaba DAMO Academy.

\subsection{Composition}

\textbf{Q: What do the instances that comprise the dataset represent (e.g., documents, photos, people, countries)?}

An instance in the dataset represents a unique programming task. Each instance consists of the following seven components. (1) Function Signature: The signature of the target code. (2) Requirement: An English description detailing the functionality of the target code. (3) Repository: The current repository contains hundreds of code files. (4) Reference Code: A developer-written implementation of the target code. This code may invoke dependencies defined in the current repository. (5) Reference Dependency: The dependencies invoked in the reference code include intra-class, intra-file, and cross-file dependencies. (6) Domain Label: The domain of the target code. (7) Test Cases: Test cases are used to check the functional correctness of the code. 

\textbf{Q: How many instances are there in total (of each type, if appropriate)?}

Our dataset is evolving and will be dynamically updated every period (\eg six months). In this paper, we release its first version - \bench-2403, which contains 275 instances.

\textbf{Q: Does the dataset contain all possible instances or is it a sample (not necessarily random) of instances from a larger set?}

Our dataset - \bench is evolving and contains a series of versions. In this paper, we release its first version - \bench-2403. In the future, we will release new versions, \eg \bench-2409. The data formats of different versions are the same.

\textbf{Q: What data does each instance consist of?}

Figure \ref{fig:Benchmark_example} shows an instance in our dataset.
Each instance consists of the following seven components. (1) Function Signature: The signature of the target code. (2) Requirement: An English description detailing the functionality of the target code. (3) Repository: The current repository contains hundreds of code files. (4) Reference Code: A developer-written implementation of the target code. This code may invoke dependencies defined in the current repository. (5) Reference Dependency: The dependencies invoked in the reference code include intra-class, intra-file, and cross-file dependencies. (6) Domain Label: The domain of the target code. (7) Test Cases: Test cases are used to check the functional correctness of the code. 

\textbf{Q: Is any information missing from individual instances?}

No.

\textbf{Q: Are relationships between individual instances made explicit (e.g., users’ movie ratings, social network links)?}

No.

\textbf{Q: Are there recommended data splits (e.g., training, development/validation, testing)?}

Our dataset is a benchmark for code generation, which only contains test data.

\textbf{Q: Are there any errors, sources of noise, or redundancies in the dataset?}

As discussed in Section \ref{sec:discussion}, we leverage LLMs to annotate requirements and domain labels of instances automatically. We conduct a human evaluation to assess the auto-generated annotations. The evaluation results (Table \ref{tab:human_evaluation}) show that auto-generated annotations are comparable to human-written annotations in most instances (\ie 96.7\% requirements and 98.5\% domain labels). In a small number of instances, auto-generated annotations may not be exactly correct, \eg missing necessary details in requirements. We think that these noises have a slight impact on our datasets. In the future, we will explore more techniques to address these noises.  

\textbf{Q: Is the dataset self-contained, or does it link to or otherwise rely on external resources (e.g., websites, tweets, other datasets)?}

It is self-contained.

\textbf{Q: Does the dataset contain data that might be considered confidential (e.g., data that is protected by legal privilege or by doctor-patient confidentiality, data that includes the content of individuals’ non-public communications)?}

No.

\textbf{Q: Does the dataset contain data that, if viewed directly, might be offensive, insulting, threatening, or might otherwise cause anxiety?}

No.

\textbf{Q: Does the dataset relate to people?} 

No.

\subsection{Collection Process}
\label{sec:app:collection}

\textbf{Q: How was the data associated with each instance acquired?}

Our dataset is collected by a four-stage pipeline, which includes (1) Repo selection and function scraping, (2) Execution-based filtering, (3) Automatic annotations, and (4) Benchmark construction. The pipeline details are in Section \ref{sec:bench:collection_pipeline}.

\textbf{Q: What mechanisms or procedures were used to collect the data (e.g., hardware apparatus or sensor, manual human curation, software program, software API)?}

Our collection pipeline involves the following four existing software tools or APIs.
\begin{itemize}
    \item GitHub APIs\footnote{\url{https://docs.github.com/en/rest?apiVersion=2022-11-28}}. We use this API to crawl open-source repositories from GitHub.
    \item Pip\footnote{\url{https://pip.pypa.io/en/stable/installation/}}. We use this software tool to install required packages for each repository automatically.
    \item Pytest\footnote{\url{https://docs.pytest.org/en/8.2.x/}}. We use this software tool to execute test cases.
    \item OpenAI API\footnote{\url{https://platform.openai.com/docs/introduction}}. Through this API, we invoke gpt-4 to generate requirements and domain labels for instances.
\end{itemize}

\textbf{Q: If the dataset is a sample from a larger set, what was the sampling strategy (e.g., deterministic, probabilistic with specific sampling probabilities)?}

N/A

\textbf{Q: Who was involved in the data collection process (e.g., students, crowdworkers, contractors) and how were they compensated (e.g., how much were crowdworkers paid)?}

N/A

\textbf{Q: Over what timeframe was the data collected?}

The first version - \bench-2403 is collected from repositories that were created between October 2023 and March 2024.

\textbf{Q: Were any ethical review processes conducted (e.g., by an institutional review board)?}

Yes. Our \bench is a code-related benchmark. It is collected from high-quality open-source repositories and excludes malicious or offensive repositories.

\subsection{Preprocessing/cleaning/labeling}

\textbf{Q: Was any preprocessing/cleaning/labeling of the data done (e.g., discretization or bucketing, tokenization, part-of-speech tagging, SIFT feature extraction, removal of instances, processing of missing values)?}

Yes. We filter out instances satisfying the following criteria: empty functions, initialization functions, and functions without executable test cases. Besides, we use a static analysis parser and an LLM to annotate these instances. Specifically, we use a static analysis parser to extract function signatures, reference code, and reference dependencies. Then, we use an LLM (gpt-4 in this paper) to annotate requirements and domain labels.

\textbf{Q: Was the “raw” data saved in addition to the preprocessed/cleaned/labeled data (e.g., to support unanticipated future uses)?}

No. Our dataset's raw data contains many large-scale code repositories, which are not conducive to downloading the dataset. We release the URLs of these raw repositories to support unanticipated future uses.

\textbf{Q: Is the software used to preprocess/clean/label the instances available?}

Yes. The used source code is available in \url{https://github.com/seketeam/EvoCodeBench}.

\subsection{Uses}

\textbf{Q: Has the dataset been used for any tasks already?}

Our dataset is designed for the code generation task. In this paper, we have evaluated eight popular code LLMs in this dataset.

\textbf{Q: Is there a repository that links to any or all papers or systems that use the dataset?}

No.

\textbf{Q: What (other) tasks could the dataset be used for?}

Besides code generation, our dataset can support the following code intelligence tasks, including code completion, test cases generation, and code summarization.

\textbf{Q: Is there anything about the composition of the dataset or the way it was collected and preprocessed/cleaned/labeled that might impact future uses?}

No.

\textbf{Q: Are there tasks for which the dataset should not be used?}

No.

\subsection{Distribution}

\textbf{Q: Will the dataset be distributed to third parties outside of the entity (e.g., company, institution, organization) on behalf of which the dataset was created?}

No.

\textbf{Q: How will the dataset will be distributed (e.g., tarball on website, API, GitHub)?}

It will be publicly available for download on GitHub (\url{https://github.com/seketeam/EvoCodeBench}).

\textbf{Q: Will the dataset be distributed under a copyright or other intellectual property (IP) license, and/or under applicable terms of use (ToU)?}

CC-4.0.

\textbf{Q: Have any third parties imposed IP-based or other restrictions on the data associated with the instances?}

No.

\textbf{Q: Do any export controls or other regulatory restrictions apply to the dataset or to individual instances?}

No.

\subsection{Maintenance}

\textbf{Q: Who is supporting/hosting/maintaining the dataset?}

The SEKE team from Peking University will continuously maintain this dataset.

\textbf{Q: How can the owner/curator/manager of the dataset be contacted (e.g., email address)?}

Please contact the first author - Jia Li (email: lijia@stu.pku.edu.cn).

\textbf{Q: Will the dataset be updated (e.g., to correct labeling errors, add new instances, delete instances)?}

Yes. We will continuously update \bench and release new versions every period (\eg six months).

\textbf{Q: Will older versions of the dataset continue to be supported/hosted/maintained?}

Yes, older versions will remain available on GitHub.

\textbf{Q: If others want to extend/augment/build on/contribute to the dataset, is there a mechanism for them to do so?}

Not officially, but our benchmark code is open source and pull requests are welcome.

\section{Collection Pipeline Details}
\label{sec:appendix:pipeline}

\subsection{Details of Automatic Annotation}
\label{sec:appendix:pipeline:annotation}

As stated in Section \ref{sec:bench:collection_pipeline}, we leverage an LLM to annotate requirements and domain labels for candidate functions.

\begin{figure}[t]
\centering
\includegraphics[width=0.8\linewidth]{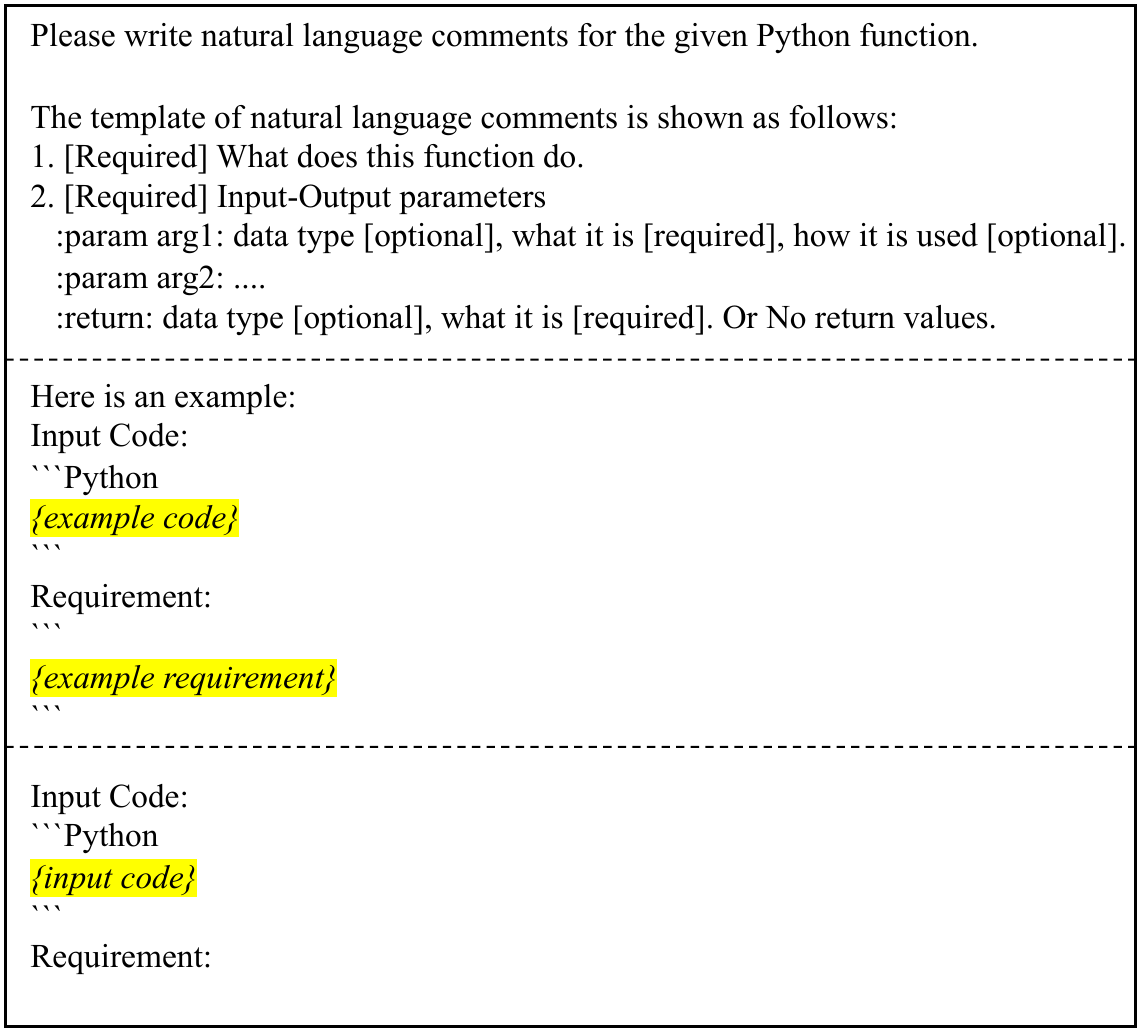}
\caption{The prompt template for generating requirements with gpt-4.}
\label{fig:gen_requirement_prompt}
\end{figure}

\begin{figure}[t]
\centering
\includegraphics[width=0.8\linewidth]{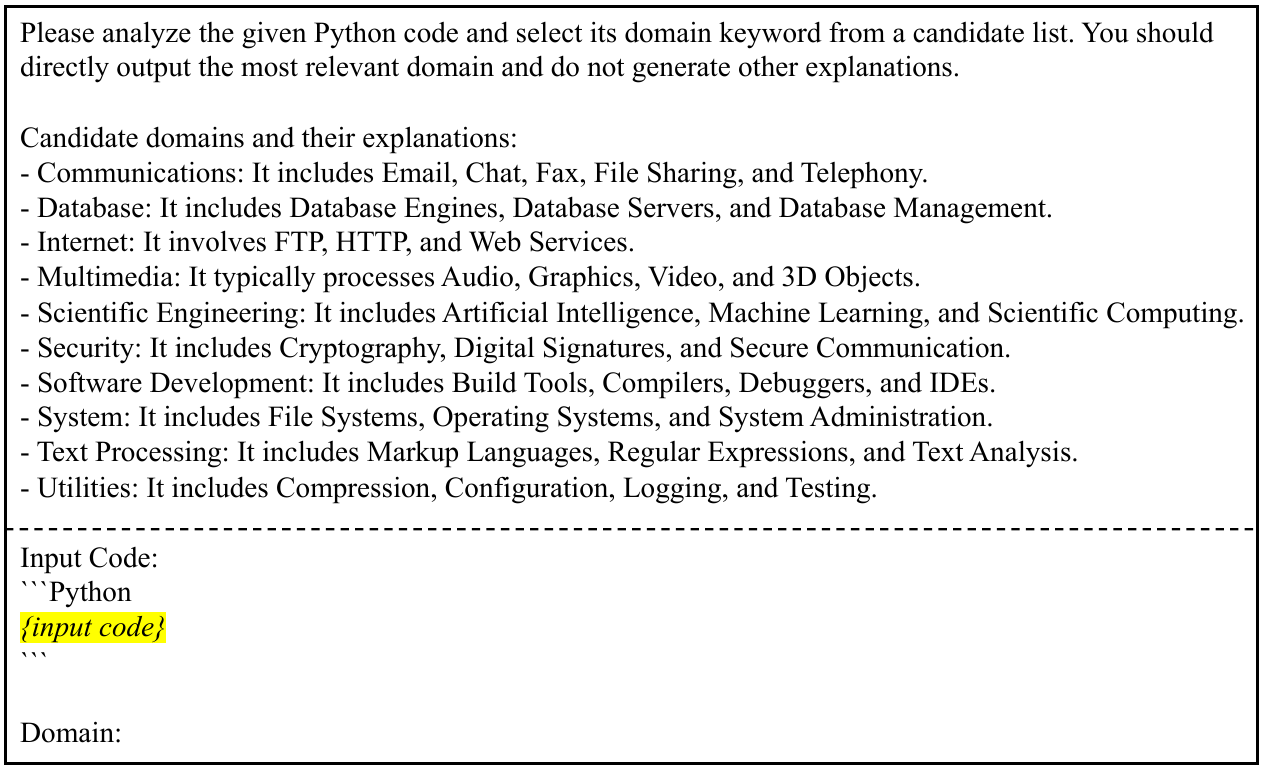}
\caption{The prompt template for generating domain labels with gpt-4.}
\label{fig:gen_domain_prompt}
\end{figure}

Figure \ref{fig:gen_requirement_prompt} and Figure \ref{fig:gen_domain_prompt} show the prompt templates for generating requirements and domain labels, respectively. The parts highlighted in yellow in the figures are placeholders. \texttt{\{example\_code\}} and \texttt{\{example\_requirement\}} are a human-written function and its requirements, respectively. We fill \texttt{\{input\_code\}} with candidate functions and leverage gpt-4 to generate requirements. We use the greedy search and generate a requirement for each function. The settings for generating domain labels are similar.

\subsection{Repositories in \bench-2403}
\label{sec:appendix:collection:projects}

\begin{table*}[t]
\centering
\caption{The statistics of 25 repositories on \bench-2403.}
\label{tab:repo_info}
\resizebox{0.8\linewidth}{!}{
\begin{tabular}{lccccc}
\toprule
Repository & Created & Stars & Py Files & Py Lines & Samples \\
\midrule
Test-Agent & 2023-10-20 & 440 & 85 & 15278 & 1  \\
skfolio & 2023-12-14 & 813 & 158 & 33852 & 13  \\
camp\_zipnerf & 2024-01-19 & 523 & 53 & 18973 & 54  \\
microagents & 2023-12-11 & 674 & 45 & 2918 & 18  \\
open-iris & 2023-12-09 & 161 & 140 & 13933 & 14  \\
litdata & 2024-02-15 & 114 & 56 & 11713 & 59  \\
nlm-ingestor & 2024-01-17 & 643 & 56 & 16674 & 4  \\
AutoRAG & 2024-01-10 & 259 & 115 & 7735 & 13  \\
XAgent & 2023-10-16 & 7054 & 148 & 17623 & 3  \\
tanuki\_py & 2023-10-16 & 606 & 108 & 10146 & 9  \\
UHGEval & 2023-11-06 & 148 & 34 & 2938 & 3  \\
Generalizable-BEV & 2023-10-30 & 136 & 570 & 132407 & 8  \\
EasyVolcap & 2023-12-07 & 442 & 308 & 51723 & 20 \\
UniRef & 2023-12-22 & 208 & 382 & 70042 & 23  \\
contrastors & 2024-01-30 & 346 & 62 & 13774 & 1  \\
gaussian-splatting-lightning & 2023-10-06 & 168 & 76 & 9935 & 1  \\
scepter & 2023-12-21 & 190 & 244 & 41519 & 1  \\
microsearch & 2024-02-05 & 336 & 5 & 231 & 2   \\
ollama-python & 2023-12-09 & 898 & 13 & 2089 & 12  \\
Python-Type-Challenges & 2023-10-23 & 343 & 121 & 3208 & 1  \\
stable-fast & 2023-10-17 & 871 & 82 & 11948 & 2  \\
stable-diffusion-webui-forge & 2024-01-14 & 2537 & 1112 & 210946 & 3  \\
openlogprobs & 2023-11-22 & 174 & 6 & 524 & 1 \\
searcharray & 2023-11-03 & 133 & 25 & 4217 & 6  \\
deluder & 2023-12-01 & 115 & 34 & 1894 & 3 \\
\bottomrule
\end{tabular}}
\end{table*}

Table \ref{tab:repo_info} shows the statistics of 25 repositories in \bench-2403.

\section{Experimental Details}
\label{sec:appendix:experiment}

\subsection{Base LLMs}
\label{sec:appendix:experiment:llm}

In this paper, we select five popular LLMs as base LLMs and evaluate them on \bench-2403. The details of these LLMs are described as follows.
\begin{itemize}
    \item \textbf{gpt-4} \cite{gpt-4}, released by OpenAI on March 14, 2023, marks another milestone in the field of natural language processing. gpt-4 demonstrates superior performance compared to previous gpt models \cite{DBLP:journals/corr/abs-2303-12712}. In our experiments, we use the version - gpt-4-1106. Its training data up to April 2023. It continues the auto-regressive prediction of the next token training objective inherited from the GPT series models. It incorporates reinforcement learning with human feedback (RLHF) and red-teaming \cite{red-teaming} techniques. However, the pre-training data scope and scale, model size, and parameters remain closed-source at present.
    
    \item \textbf{gpt-3.5-turbo} \cite{gpt-3.5} is an improved gpt-3 model enhanced by a three-stage reinforcement learning with human feedback (RLHF) algorithm. Apart from improving instruction-following capabilities, the RLHF algorithm proves highly effective in mitigating the generation of harmful or toxic content, which is crucial for the practical deployment of LLMs in security-sensitive contexts. we utilized the released versions of gpt-3.5, namely gpt-3.5-turbo-1106, with training data up to September 2021. However, similar to gpt-4, the training details, training data, and model weights are currently closed-source.  

    \item \textbf{CodeLLaMa} \cite{CodeLLaMa}, based on the LLama2 architecture by Meta-AI\footnote{\url{https://ai.meta.com/}}, was fine-tuned and open-sourced by the company on August 25, 2023, with versions of 7B, 13B, and 34B. A 70B version was released on January 30, 2024 \cite{CodeLLaMa}. CodeLLama is primarily trained on nearly deduplicated publicly available code datasets. The first three models were trained on 500 billion tokenized code, while the latest 70B model was trained on 1T tokens. Similar to the LLaMa series, CodeLLaMa also follows a decoder-only architecture. We evaluated CodeLLaMa-Python-\{7B, 13B\} upon our \bench.
       
    \item \textbf{DeepSeek Coder} \cite{DeepSeek_Coder} is a large language model for programming tasks released by DeepSeek-AI\footnote{\url{https://www.deepseek.com/}} in November 2, 2023. DeepSeek Coder consists of a series of code language models, each trained from scratch on 2T tokens, containing 87\% code and 13\% natural language. DeepSeek Coder provides code models with 1.3B, 6.7B and 33B parameter sizes. In terms of model architecture, each model integrates a decoder-only Transformer, incorporating Rotary Position Embedding and FlashAttention v2. We evaluated DeepSeek Coder-\{6.7B, 33B\} on our \bench.

    \item  \textbf{StarCoder 2} \cite{StarCoder-2} was released by BigCode\footnote{\url{https://www.bigcode-project.org/}} on December 8, 2023 with 3 different parameters, 3B, 7B and 15B. StarCoder2 is trained on The Stack v2, a new large-scale, high-quality code dataset. All models were trained using Grouped Query Attention, a contextual window of 16,384 tokens with a sliding window attention of 4,096 tokens, using the Fill-in-the-Middle objective. Following DeepseekCoder \cite{DeepSeek_Coder} and Code LLaMA \cite{CodeLLaMa}, StarCoder2 use Rotary Positional Encodings. We evaluated StarCoder2-\{7B, 15B\} on our \bench, which was trained on over 3.5 trillion tokens in 17 programming languages from Stack v2.


    
\end{itemize}

\subsection{Prompt Templates}
\label{sec:appendix:experiment:prompt}

\begin{figure}[t]
\centering
\includegraphics[width=0.8\linewidth]{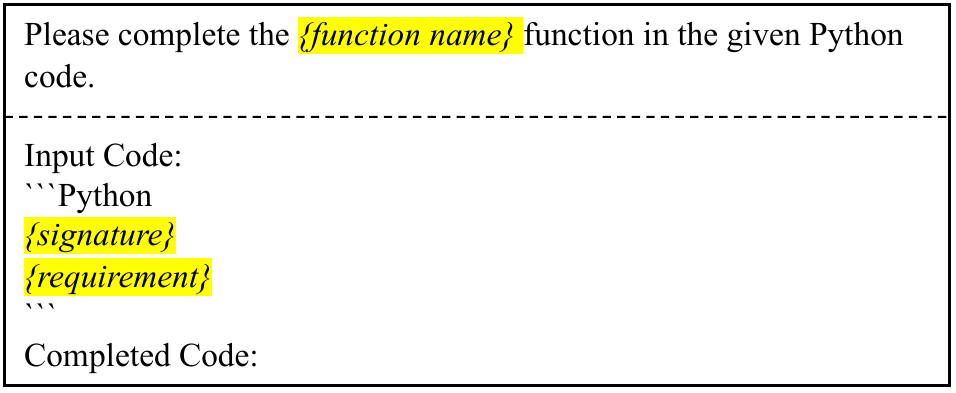}
\caption{The prompt template in the without context setting.}
\label{fig:nl_prompt}
\end{figure}

\begin{figure}[t]
\centering
\includegraphics[width=0.8\linewidth]{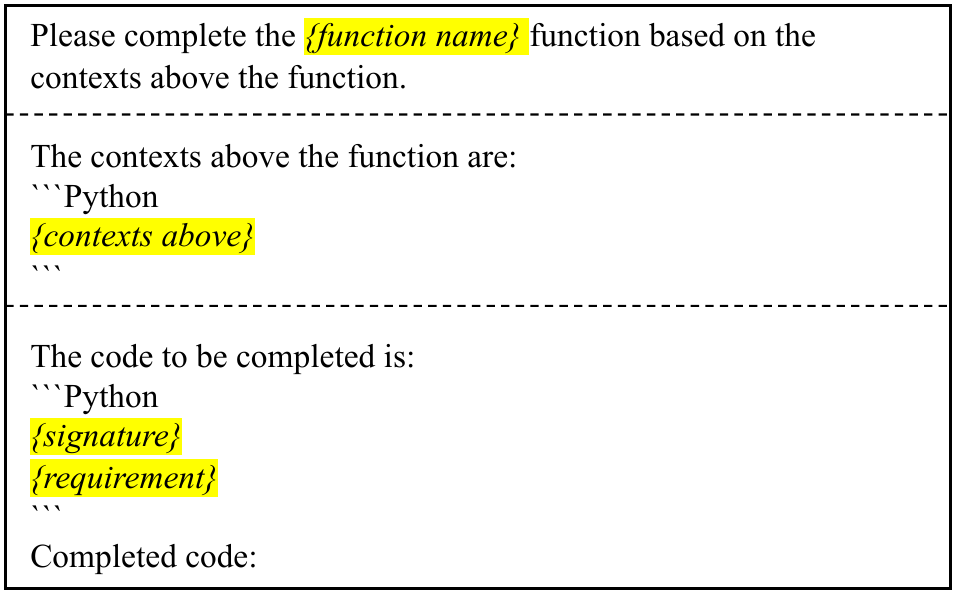}
\caption{The prompt template in the local file (completion) setting.}
\label{fig:local_completion_prompt}
\end{figure}

\begin{figure}[t]
\centering
\includegraphics[width=0.8\linewidth]{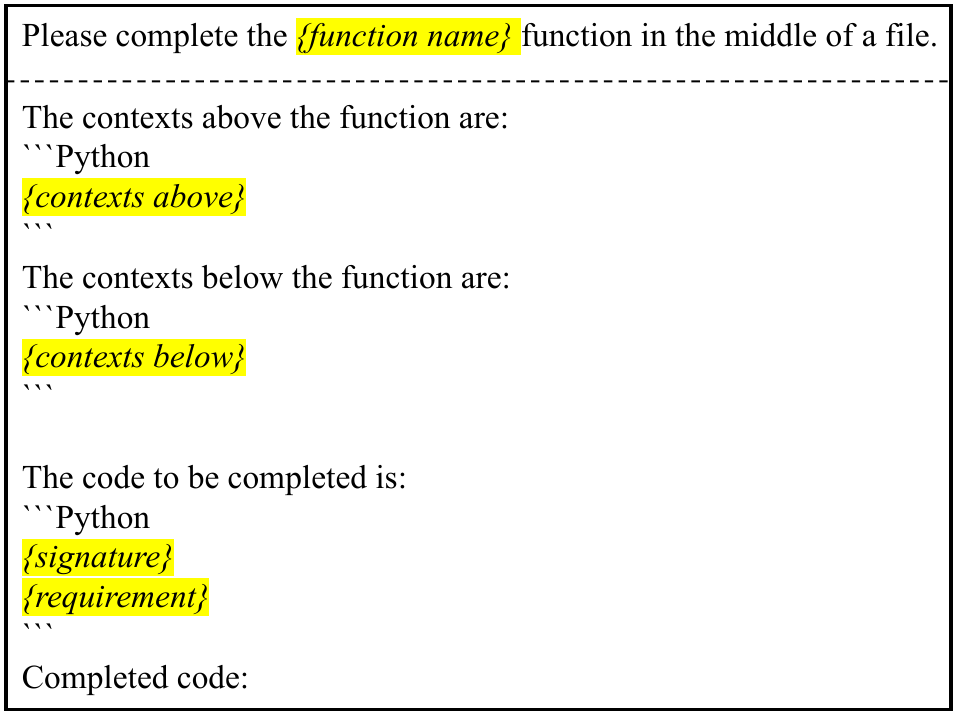}
\caption{The prompt template in the local file (infilling) setting.}
\label{fig:local_infilling_prompt}
\end{figure}

\begin{figure}[t]
\centering
\includegraphics[width=0.8\linewidth]{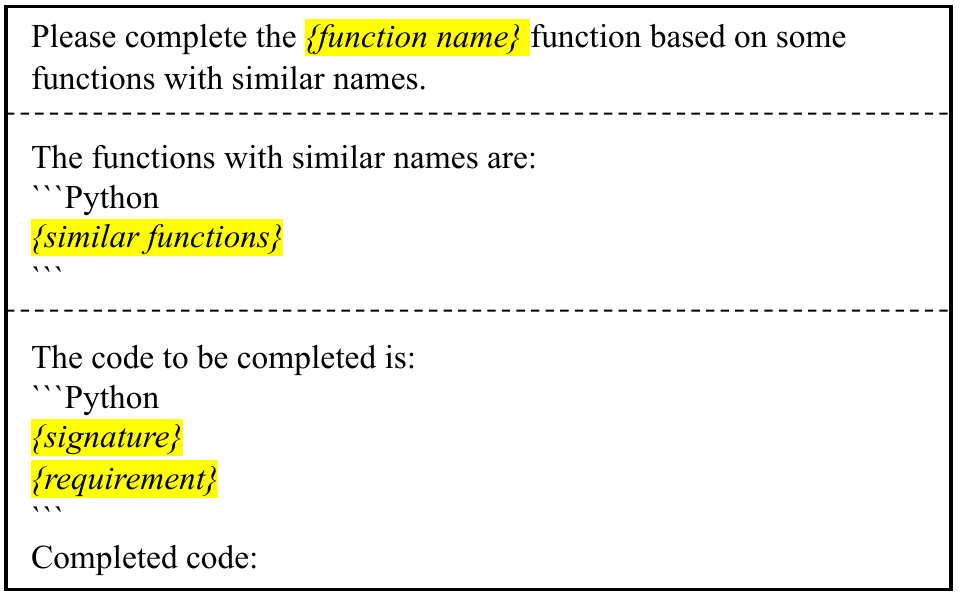}
\caption{The prompt template in the similar function setting.}
\label{fig:similar_func_prompt}
\end{figure}

The prompt templates used for instruction-tuning models (\ie gpt-4 and gpt-3.5) are shown in Figure \ref{fig:nl_prompt}, \ref{fig:local_completion_prompt}, \ref{fig:local_infilling_prompt}, and \ref{fig:similar_func_prompt}. \texttt{\{function name\}}, \texttt{\{contexts above\}}, \texttt{\{contexts below\}}, \texttt{\{signature\}}, and \texttt{\{requirement\}} are placeholders.

For other standard language models, the prompt templates are shown as follows: 
\begin{itemize}
    \item Without context: [\texttt{signature; requirement}]
    \item Local file (completion): [\texttt{context\_above; signature; requirement}]
    \item Local file (infilling): [\texttt{prefix\_id; context\_above; signature; requirement; suffix\_id; context\_below; middle\_id}]
\end{itemize}
Where \texttt{[;]} denotes the concatenation operation of strings. \texttt{\{prefix\_id\}}, \texttt{\{suffix\_id\}}, \texttt{\{middle\_id\}} are special tokens used in code infilling. For different LLMs, we reuse their official special tokens to make prompts. 

\subsection{Details of Human Evaluation}
\label{sec:appendix:experiment:human_eval}

\begin{figure}[t]
\centering
\includegraphics[width=0.8\linewidth]{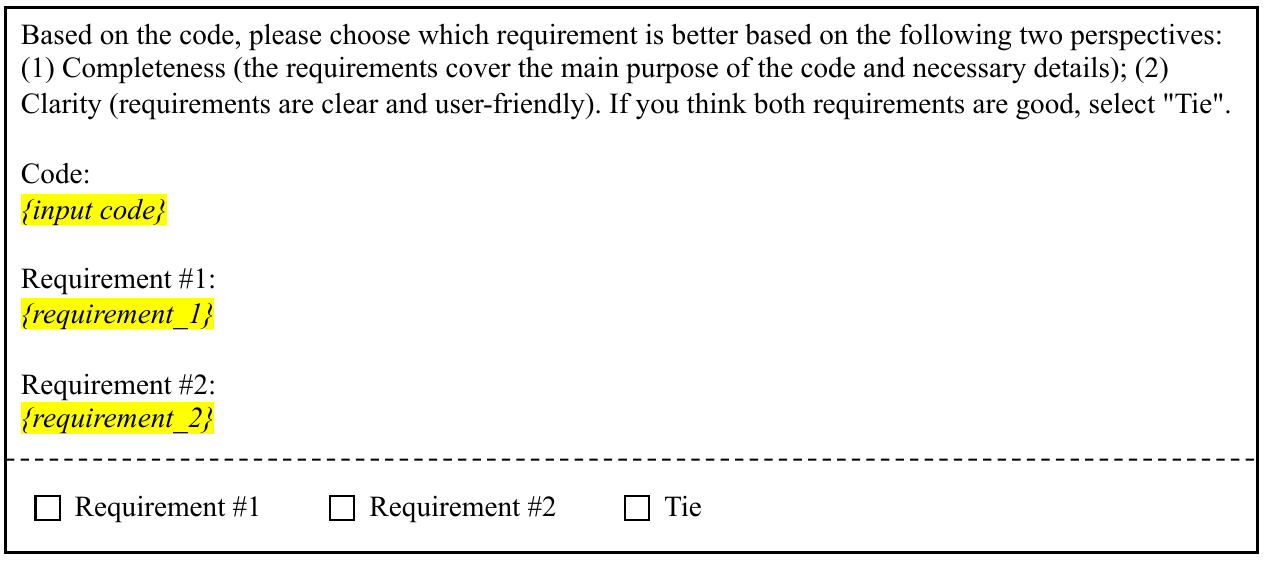}
\caption{The questionnaire template for evaluating requirements.}
\label{fig:requirement_evaluation}
\end{figure}

\begin{figure}[t]
\centering
\includegraphics[width=0.8\linewidth]{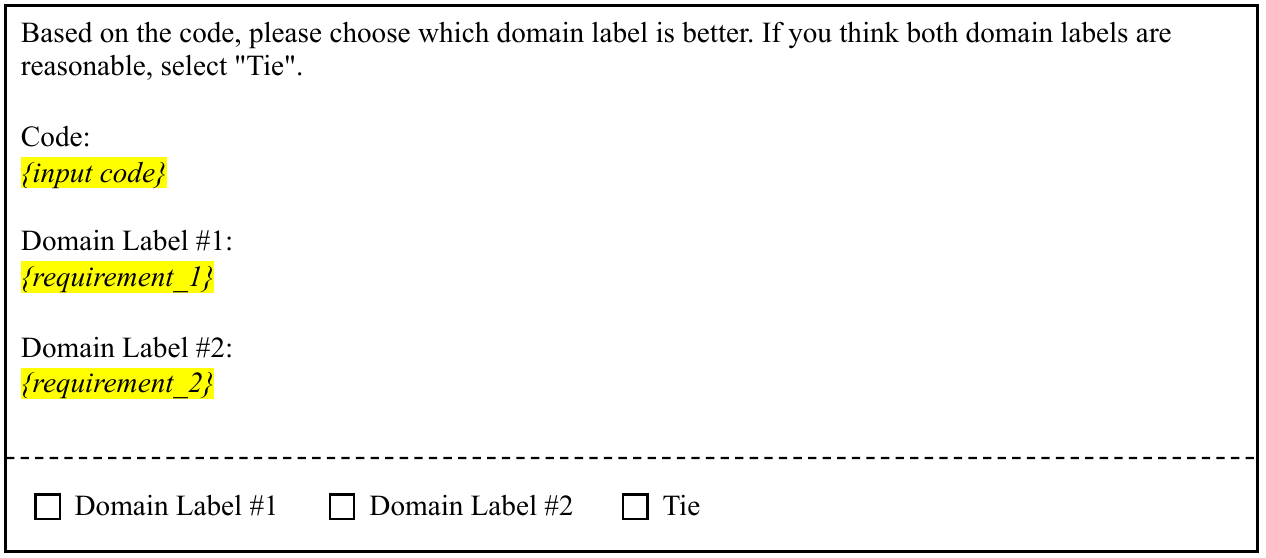}
\caption{The questionnaire template for evaluating domain labels.}
\label{fig:domain_evaluation}
\end{figure}

Figure \ref{fig:requirement_evaluation} and Figure \ref{fig:domain_evaluation} show the questionnaire templates for evaluating requirements and domain labels, respectively. The parts highlighted in yellow in the figures are placeholders. Taking Figure \ref{fig:requirement_evaluation} as an example, we randomly arrange auto-generated requirements and human-written requirements and then fill them into the placeholders. The setup for evaluating domain labels is similar.

\end{document}